\RequirePackage{fix-cm}

\documentclass[twocolumn]{svjour3}

\usepackage{array}
\usepackage{hhline}
\usepackage{multirow}
\usepackage{color}
\usepackage{amsmath}
\usepackage{amssymb}
\usepackage{graphics}
\usepackage{graphicx}
\usepackage{epstopdf}
\usepackage{epsfig}
\usepackage{import}

\usepackage{soul}

\usepackage[pagebackref=true,breaklinks=false,letterpaper=true,colorlinks,bookmarks=false]{hyperref}
\usepackage{breakurl}
\usepackage{ifthen}
\newboolean{ABR}
\setboolean{ABR}{false}
\newboolean{FTR}
\setboolean{FTR}{false}

\def\Div{\textup{div\,}}
\def\*#1{\mathbf{#1}}

\newcommand{\off}[1]{}

\newenvironment{itemizepoint}{%
 \begin{itemize}}{\end{itemize}}

\title{Normal Integration: A Survey}

\author{Yvain Qu\'eau \and Jean-Denis Durou \and Jean-Fran\c cois Aujol}

\institute{Y. Qu\'eau \at
	Technical University Munich, Germany \\
	\email{yvain.queau@tum.de}
\and
	J.-D. Durou \at
	IRIT, Universit\'e de Toulouse, France
\and
	J.-F. Aujol \at
	IMB, Universit\'e de Bordeaux, France \\
	Institut Universitaire de France
}

\begin{document}

\maketitle

\begin{abstract}
The need for efficient normal integration me\-thods is driven by several computer vision tasks such as shape-from-shading, photometric stereo, deflectometry, etc. In the first part of this survey, we select the most important properties that one may expect from a normal integration method, based on a thorough study of two pioneering works by Horn and Brooks \cite{Horn:1986a} and by Frankot and Chellappa \cite{Frankot:1988a}. Apart from accuracy, an integration method should at least be fast and robust to a noisy normal field. In addition, it should be able to handle several types of boundary condition, including the case of a free boundary, and a reconstruction domain of any shape i.e., which is not necessarily rectangular. It is also much appreciated that a minimum number of parameters have to be tuned, or even no parameter at all. Finally, it should preserve the depth discontinuities. In the second part of this survey, we review most of the existing methods in view of this analysis, and conclude that none of them satisfies all of the required properties. This work is complemented by a companion paper entitled \emph{Variational Methods for Normal Integration}, in which we focus on the problem of normal integration in the presence of depth discontinuities, a problem which occurs as soon as there are occlusions.
\end{abstract}



\section{Introduction}

Computing the 3D-shape of a surface from a set of normals is a classical problem of 3D-reconstruction called \emph{normal integration}. This problem is well-posed, except that a constant of integration has to be fixed, but its resolution is not as straightforward as it could appear, even in the case where the normal is known at every pixel of an image. One may well be surprised that such a simple problem has given rise to such a large number of papers. This is probably due to the fact that, like many computer vision problems, it simultaneously meets several requirements. Of course, a method of integration is expected to be accurate, fast, and robust re noisy data or outliers, but we will see that several other criteria are important as well.

In this paper, a thorough study of two pioneering works is done: a paper by Horn and Brooks based on variational calculus \cite{Horn:1986a}; another one by Frankot and Chellappa resorting to Fourier analysis \cite{Frankot:1988a}. This preliminary study allows us to select six criteria apart from accuracy, through which we intend to \emph{qualitatively} evaluate the main normal integration methods. Our survey is summarized in Table \ref{tab:1}. Knowing that no existing method is completely satisfactory, this preliminary study impels us to suggest several new methods of integration which will be found in a companion paper \cite{Queau:2017}.

The organization of the present paper is the following. We derive the basic equations of normal integration in Section \ref{sec:2}. Horn and Brooks' and Frankot and Chellappa's methods are reviewed in Section \ref{sec:3}. This allows us, in Section \ref{sec:4}, to select several properties that are required by any normal integration method, and to comment the most relevant related works. In Section \ref{sec:5}, we conclude that a completely satisfactory method of integration is still lacking.


\section{Basic Equations of Normal Integration} \label{sec:2}

Suppose that, in each point $\overline{\*x} = [u,v]^\top$ of the image of a surface, the outer unit-length normal $\*n(u,v) = [n_1(u,v), n_2(u,v), n_3(u,v)]^\top$ is known. Integrating the \emph{normal field} $\*n$ amounts to searching for three functions $x$, $y$ and $z$ such that the normal to the surface at the surface point $\*x(u,v) = [x(u,v), y(u,v), z(u,v)]^\top$, which is conjugate to $\overline{\*x}$, is equal to $\*n(u,v)$. Let us rigorously formulate this problem when the projection model is either orthographic, weak-perspective or perspective.


\subsection{Orthographic Projection}

\begin{figure}[htb]
\begin{center}
	\def\svgwidth{0.5 \textwidth}
		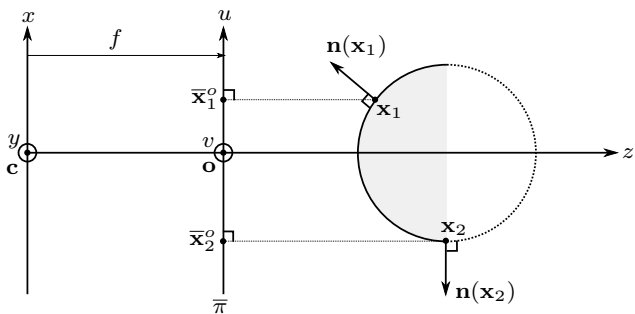
\end{center}
\caption{Orthographic projection: $\*x_1$ and $\*x_2$ are conjugate to $\overline{\*x}^o_1$ and $\overline{\*x}^o_2$, respectively. The visible part of the surface is highlighted in gray. Point $\overline{\*x}^o_2$ lies on the occluding contour.}
\label{fig:1}
\end{figure}

We attach to the camera a 3D-frame $\*cxyz$ whose origin~$\*c$ is located at the optical center, and such that $\*cz$ coincides with the optical axis (see Figure \ref{fig:1}).

The origin of pixel coordinates is taken as the intersection $\*o$ of the optical axis $\*cz$ and the image plane~$\overline{\pi}$. In practice, $\overline{\pi}$ coincides with the focal plane $z = f$, where $f$ denotes the focal length.

Assuming orthographic projection, a 3D-point $\*x$ pro\-jects orthogonally onto the image plane, i.e.:
\begin{equation}
	\left\{
	\begin{aligned}
		x(u,v) &= u \\
		y(u,v) &= v
	\end{aligned}
	\right.
\label{eq:1}
\end{equation}
By normalizing the cross product of both partial derivatives $\partial_u \*x$ and $\partial_v \*x$, and choosing the sign so that $\*n$ points towards the camera, we obtain (the dependencies in $u$ and $v$ are omitted, for the sake of simplicity):
\begin{equation}
	\*n = \frac{1}{\sqrt{1+\|\nabla{z}\|^2}}
	\left[
		\begin{array}{c}
			\partial_u{z} \\
			\partial_v{z} \\
			-1
		\end{array}
	\right]
\label{eq:2}
\end{equation}
where $\nabla z = [\partial_u z,\partial_v z]^\top$ denotes the gradient of the \emph{depth map} $z$. From \eqref{eq:2}, we conclude that $n_3 < 0$.

Of course, 3D-points such that $n_3 \geq 0$ also exist. Such points are non-visible if $n_3 > 0$. If $n_3 = 0$, they are visible and project onto the \emph{occluding contour} (see point $\overline{\*x}^o_2$ in Figure \ref{fig:1}). Thus, even if the normal $\*n$ is easily determined on the occluding contour, since $\*n$ is both parallel to $\overline{\pi}$, and orthogonal to the contour, computing the depth $z$ by integration in such points is impossible.

Now, let us consider the image points which do not lie on the occluding contour. Equation \eqref{eq:2} immediately gives the following pair of linear PDEs in $z$:
\begin{equation}
	\nabla z = [p, q]^\top
\label{eq:3}
\end{equation}
where:
\begin{equation}
	\left\{
	\begin{aligned}
		p = -\frac{n_1}{n_3} \\
		q = -\frac{n_2}{n_3}
	\end{aligned}
	\right.
\label{eq:4}
\end{equation}
Equations \eqref{eq:3} show that integrating a normal field i.e., computing $z$ from $\*n$, amounts to integrating the vector field $[p,q]^\top$. The solution of \eqref{eq:3} is straightforward:
\begin{equation}
	z(u,v) = z(u_0,v_0)+\int_{(r,s)=(u_0,v_0)}^{(u,v)} \!\!\!\!\!\!\!\!\!\!\!\!\!\!\!\!\!\! \left[p(r,s)\,\mathrm{d}r+q(r,s)\,\mathrm{d}s\right]
\label{eq:5}
\end{equation}
regardless of the integration path between some point $(u_0,v_0)$ and $(u,v)$, as soon as $p$ and $q$ satisfy the \emph{constraint of integrability} $\partial_v p = \partial_u q$ (Schwartz theorem). If they do not, the integral in Equation \eqref{eq:5} depends on the integration path.

If there is no point $(u_0,v_0)$ where $z$ is known, it follows from \eqref{eq:5} that $z(u,v)$ is computable up to an additive constant. This constant can be chosen so as to minimize the root mean square error (RMSE) in $z$, provided that the ground-truth is available\footnote{The same procedure is used by Klette and Schl{\"u}ns in \cite{Klette:1996a}: ``The reconstructed height values are shifted in the range of the original surface using LSE optimization''.}.


\subsection{Weak-perspective Projection}

\begin{figure}[htb]
\begin{center}
	\def\svgwidth{0.5 \textwidth}
		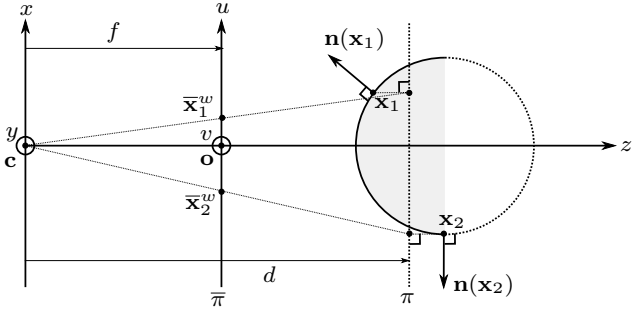
\end{center}
\caption{Weak-perspective projection: $\*x_1$ and $\*x_2$ are conjugate to $\overline{\*x}_1^w$ and $\overline{\*x}_2^w$, respectively. Point $\overline{\*x}_2^w$ lies on the occluding contour. The plane $\pi$, which is conjugate to $\overline{\pi}$, is supposed to match the mean location of the surface.}
\label{fig:2}
\end{figure}

Weak-perspective projection assumes that the camera is focused on a plane $\pi$ of equation $z = \mathsf{d}$, supposed to match the mean location of the surface. Any visible point $\*x$ projects first orthogonally onto $\pi$, then perspectively onto $\overline{\pi}$, with $\*c$ as projection center (see Figure \ref{fig:2}).

Assuming weak-perspective projection, we have:
\begin{equation}
	\left\{
	\begin{aligned}
		x(u,v) &= \displaystyle\frac{\mathsf{d}}{\mathsf{f}}\,u \\
		y(u,v) &= \displaystyle\frac{\mathsf{d}}{\mathsf{f}}\,v
	\end{aligned}
	\right.
\label{eq:6}
\end{equation}

Denoting by $\mathsf{m}$ the \emph{image magnification} $\mathsf{f}/\mathsf{d}$, the outer unit-length normal $\*n$ now reads:
\begin{equation}
	\*n = \frac{1}{\sqrt{1+\mathsf{m}^2\|\nabla{z}\|^2}}
	\left[
		\begin{array}{c}
			\mathsf{m}\,\partial_u{z} \\
			\mathsf{m}\,\partial_v{z} \\
			-1
		\end{array}
	\right]
\label{eq:7}
\end{equation}

\indent{For the image points which do not lie on the occluding contour, the pair of PDEs \eqref{eq:3} becomes:
\begin{equation}
	\nabla z = \frac{1}{\mathsf{m}}\,[p, q]^\top
\label{eq:8}
\end{equation}
which explains why ``weak-perspective projection'' is also called ``scaled orthographic projection''. From \eqref{eq:8}, we easily extend \eqref{eq:5}:
\begin{equation}
	z(u,v) = z(u_0,v_0)+\frac{1}{\mathsf{m}} \!\int_{(r,s)=(u_0,v_0)}^{(u,v)} \!\!\!\!\!\!\!\!\!\!\!\!\!\!\!\!\!\! \left[p(r,s)\,\mathrm{d}r+q(r,s)\,\mathrm{d}s\right]
\label{eq:9}
\end{equation}


\subsection{Perspective Projection}

We now consider perspective projection (see Figure \ref{fig:3}).
\begin{figure}[htb]
\begin{center}
	\def\svgwidth{0.5 \textwidth}
		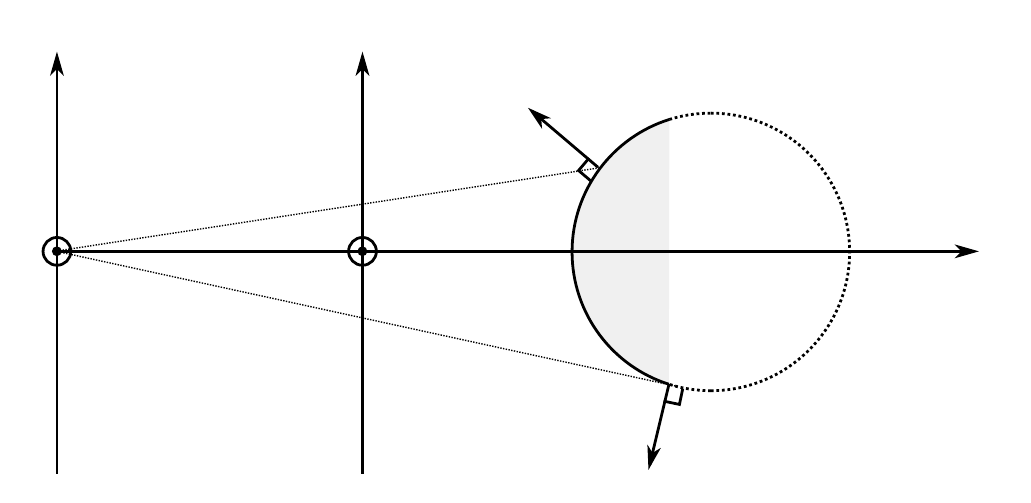
\end{center}
\caption{Perspective projection: $\*x_1$ and $\*x_3$ are conjugate to $\overline{\*x}_1^p$ and $\overline{\*x}_3^p$, respectively ($\overline{\*x}_3^p$ lies on the occluding contour).}
\label{fig:3}
\end{figure}

As major difference with weak-perspective, $\mathsf{d}$ must be replaced with $z(u,v)$ in Equations \eqref{eq:6}:
\begin{equation}
	\left\{
	\begin{aligned}
		x(u,v) = \displaystyle\frac{z(u,v)}{\mathsf{f}}\,u \\
		y(u,v) = \displaystyle\frac{z(u,v)}{\mathsf{f}}\,v
	\end{aligned}
	\right.
\label{eq:10}
\end{equation}
The cross product of $\partial_u{\*x}$ and $\partial_v{\*x}$ is a little more complicated than in the previous case:
\begin{equation}
	\partial_u{\*x_p} \times \partial_v{\*x_p} = \frac{z}{\mathsf{f}^2}\,
	\left[
		\begin{array}{c}
			-\mathsf{f}\,\partial_u z \\
			-\mathsf{f}\,\partial_v z \\
			z+u\,\partial_u z+v\,\partial_v z
		\end{array}
	\right]
\label{eq:11}
\end{equation}
Writing that this vector is parallel to $\*n(u,v)$, this provides us with the three following equations:
\begin{equation}
	\left\{
	\begin{aligned}
		& \mathsf{f}\,n_3\,\partial_u z+n_1\,[z+u\,\partial_u z+v\,\partial_v z] = 0 \\
		& \mathsf{f}\,n_3\,\partial_v z+n_2\,[z+u\,\partial_u z+v\,\partial_v z] = 0 \\
		& n_2\,\partial_u z-n_1\,\partial_v z = 0
	\end{aligned}
	\right.
\label{eq:12}
\end{equation}
Since this system is homogeneous in $z$, and knowing that $z>0$, we introduce the change of variable:
\begin{equation}
	\tilde{z} = \ln(z)
\label{eq:13}
\end{equation}
which makes \eqref{eq:12} linear with respect to $\partial_u \tilde{z}$ and $\partial_v \tilde{z}$:
\begin{equation}
	\left\{
	\begin{aligned}
		& [\mathsf{f}\,n_3+u\,n_1]\, \partial_u \tilde{z}+v\,n_1\, \partial_v \tilde{z} = -n_1 \\
		& u\,n_2\, \partial_u \tilde{z}+[\mathsf{f}\,n_3+v\,n_2]\, \partial_v \tilde{z} = -n_2 \\
		& n_2\, \partial_u \tilde{z}-n_1\, \partial_v \tilde{z} = 0
	\end{aligned}
	\right.
\label{eq:14}
\end{equation}
System \eqref{eq:14} is non-invertible if it has rank less than $2$ i.e., if its three determinants are zero:
\begin{equation}
	\left\{
	\begin{aligned}
		\mathsf{f}\,n_3\,[u\,n_1+v\,n_2+\mathsf{f}\,n_3] = 0 \\
		-n_1\,[u\,n_1+v\,n_2+\mathsf{f}\,n_3] = 0 \\
		-n_2\,[u\,n_1+v\,n_2+\mathsf{f}\,n_3] = 0
	\end{aligned}
	\right.
\label{eq:15}
\end{equation}
As $n_1$, $n_2$ and $n_3$ cannot simultaneously vanish, because vector $\*n$ is unit-length, the equalities \eqref{eq:15} holds true if and only if $u\,n_1+v\,n_2+\mathsf{f}\,n_3 = 0$. Knowing that $[u,v,\mathsf{f}]^\top$ are the coordinates of the image point $\overline{\*x}^p$ in the 3D-frame $\*cxyz$, this happens if and only if $\overline{\*x}^p$ lies on the occluding contour (see point $\overline{\*x}_3^p$ in Figure \ref{fig:3}). It is noticeable that for a given object, the occluding contour depends on which projection model is assumed.

For the image points which do not lie on the occluding contour, System \eqref{eq:14} is easily inverted, which gives us the following pair of linear PDEs in $\tilde{z}$:
\begin{equation}
	\nabla \tilde{z} = [\tilde{p}, \tilde{q}]^\top
\label{eq:16}
\end{equation}
where:
\begin{equation}
	\left\{
	\begin{aligned}
		\tilde{p} &= -\displaystyle\frac{n_1}{u\,n_1+v\,n_2+\mathsf{f}\,n_3} \\
		\tilde{q} &= -\displaystyle\frac{n_2}{u\,n_1+v\,n_2+\mathsf{f}\,n_3}
	\end{aligned}
	\right.
\label{eq:17}
\end{equation}

\noindent{Hence, under perspective projection, integrating a normal field~$\*n$ amounts to integrating the vector field $[\tilde{p}, \tilde{q}]^\top$. The solution of Equation~\eqref{eq:16} is straightforward:}
\begin{equation}
	\tilde{z}(u,v) = \tilde{z}(u_0,v_0)+\int_{(r,s)=(u_0,v_0)}^{(u,v)}\!\!\!\!\!\!\!\!\!\!\!\!\left[\tilde{p}(r,s)\,\mathrm{d}r+\tilde{q}(r,s)\,\mathrm{d}s\right]
\label{eq:18}
\end{equation}
from which we deduce, using \eqref{eq:13}:
\begin{equation}
	z(u,v) = z(u_0,v_0)\,\exp{\left\{\int_{(r,s)=(u_0,v_0)}^{(u,v)}\!\!\!\!\!\!\!\!\!\!\!\!\!\!\!\left[\tilde{p}(r,s)\,\mathrm{d}r+\tilde{q}(r,s)\,\mathrm{d}s\right]\right\}}
\label{eq:19}
\end{equation}

It follows from \eqref{eq:19} that $z$ is computable up to a multiplicative constant. Notice also that \eqref{eq:17}, and therefore \eqref{eq:19}, require that the focal length $\mathsf{f}$ is known, as well as the location of the principal point $\*o$, since the coordinates $u$ and $v$ depend on it (see Figure \ref{fig:3}).

The similarity between \eqref{eq:3}, \eqref{eq:8} and \eqref{eq:16} shows that any normal integration method can be extended to weak-perspective or perspective, provided that the intrinsic parameters of the camera are known\footnote{This is advocated in \cite{Tankus:2005b} for the method designed in \cite{Horovitz:2004a}.}. Let us emphasize that such extensions are \emph{generic} i.e., not restricted to a given method of integration. We can thus limit ourselves to solving the following pair of linear PDEs, which we consider as the \emph{model problem}:
\begin{equation}
	\nabla z = \mathbf{g}
\label{eq:20}
\end{equation}
where $(z,\mathbf{g})$ means $(z,[p,q]^\top)$, $(z,\frac{1}{\mathsf{m}}\,[p,q]^\top)$, or $(\tilde{z},[\tilde{p},\tilde{q}]^\top)$, depending on whether the projection model is orthographic, weak-perspective, or perspective, respectively.


\subsection{Integration Using Quadratic Regularization} \label{sec:2.3}

From now on, we do not care more about the projection model. We just have to solve the generic equation \eqref{eq:20}.

As already noticed, the respective solutions \eqref{eq:5}, \eqref{eq:9} and \eqref{eq:18} of Equations \eqref{eq:3}, \eqref{eq:8} and \eqref{eq:16}, are independent from the integration path if and only if the constraint of integrability $\partial_v p = \partial_u q$, or $\partial_v \tilde{p} = \partial_u \tilde{q}$, is satisfied. In practice, a normal field is never rigorously integrable (or \emph{curl-free}). Apart from using several integration paths and meaning the integrals \cite{Coleman:1982a,Healey:1984a,Wu_Z:1988a}, a natural way to deal with the lack of integrability is to turn \eqref{eq:20} into an optimization problem \cite{Horn:1986a}. Using quadratic regularization, this amounts to minimizing the functional:
\begin{equation}
	\mathcal{F}_{L_2}(z) = \iint\limits_{(u,v) \in \Omega} \|\nabla z(u,v)-\*g(u,v)\|^2 \,\mathrm{d}u\,\mathrm{d}v
\label{eq:21}
\end{equation}
where $\Omega \subset \mathbb{R}^2$ is the \emph{reconstruction domain}, and the \emph{gradient field} $\*g = [p, q]^\top$ is the datum of the problem.

The functional $\mathcal{F}_{L_2}$ is strictly convex in $\nabla z$, but does not admit a unique minimizer $z^*$ since, for any $\kappa\in\mathbb{R}$, $\mathcal{F}_{L_2}(z^*+\kappa) = \mathcal{F}_{L_2}(z^*)$. Its minimization requires that the associated Euler-Lagrange equation is satisfied. The calculus of variation provides us with the following:
\begin{equation}
	\nabla \mathcal{F}_{L_2}(z) = 0 \quad
	\Longleftrightarrow \quad
	-2 \, \Div \left( \nabla z - \*g \right) = 0
\label{eq:22}
\end{equation}
This \emph{necessary} condition is rewritten as the following \emph{Poisson equation}\footnote{Similar equations arise in other computer vision problems \cite{Aubert:2014,Perez:2003,Simchony:1990a}.}:
\begin{equation}
	\Delta z = \partial_u p + \partial_v q
\label{eq:23}
\end{equation}

Solving Equation \eqref{eq:23} is not a \emph{sufficient} condition for minimizing $\mathcal{F}_{L_2}(z)$, except if $z$ is known on the boundary $\partial\Omega$ (Dirichlet boundary condition), see \cite{Miranda} and the references therein. Otherwise, the so-called \emph{natural boundary condition}, which is of the Neumann type, must be considered. In the case of $\mathcal{F}_{L_2}(z)$, this condition is written \cite{Horn:1986a}:
\begin{equation}
	(\nabla z -\*g) \cdot \boldsymbol{\eta} = 0
\label{eq:24}
\end{equation}
where vector $\boldsymbol{\eta}$ is normal to $\partial\Omega$ in the image plane.

Using different boundary conditions, one could expect that the different solutions of \eqref{eq:23} would coincide on most part of $\Omega$, but this is not true. For a given gradient field $\*g$, the choice of a boundary condition has a great influence on the recovered surface. This is noted in \cite{Horn:1986a}: ``Equation [\eqref{eq:23}] does not uniquely specify a solution without further constraint. In fact, we can add any harmonic function to a solution to obtain a different solution also satisfying [\eqref{eq:23}]''. A \emph{harmonic function} is a solution of the \emph{Laplace equation}:
\begin{equation}
	\Delta z = 0
\label{eq:25}
\end{equation}
As an example, let us search for the harmonic functions taking the form $z(u,v) = z_1(u)\,z_2(v)$. Knowing that $z_1\neq0$ and $z_2\neq0$, since $z > 0$, Equation \eqref{eq:25} gives:
\begin{equation}
	\frac{z_1''(u)}{z_1(u)} = -\frac{z_2''(v)}{z_2(v)}
\label{eq:26}
\end{equation}
Both sides of Equation \eqref{eq:26} are thus equal to the same constant $K\in\mathbb{R}$. Two cases may occur, according to the sign of $K$. If $K<0$, we pose $K = -\omega^2$, $\omega\in\mathbb{R}$:
\begin{equation}
	\left\{
		\begin{array}{l}
			z_1''(u)+\omega^2\,z_1(u) = 0 \\
			z_2''(v)-\omega^2\,z_2(v) = 0
		\end{array}
	\right. \,
	\Rightarrow \,
	\left\{
		\begin{array}{l}
			z_1(u) = z_1(0)\,e^{\mathsf{j}\,\omega u} \\
			z_2(v) = z_2(0)\,e^{\omega v}
		\end{array}
	\right.
\label{eq:27}
\end{equation}
where $\mathsf{j}$ is such that $\mathsf{j}^2 = -1$. Finally, we obtain:
\begin{equation}
	z(u,v) = z(0,0)\,e^{\omega(\mathsf{j}\,u+v)}
\label{eq:28}
\end{equation}}
\indent{Note that a real harmonic function defined on $\mathbb{R}^2$ can be considered as the real part or as the imaginary part of a \emph{holomorphic function}. All the functions of the form \eqref{eq:28} are indeed holomorphic. Their real and imaginary parts thus provide us with the following two families of harmonic functions:
\begin{equation}
	\left\{\cos(\omega u)\,e^{\omega v}\right\}_{\omega\in\mathbb{R}} \quad \text{;} \quad
	\left\{\sin(\omega u)\,e^{\omega v}\right\}_{\omega\in\mathbb{R}}
\label{eq:29}
\end{equation}
Adding to a given solution of Equation \eqref{eq:23} any linear combination of these harmonic functions (many other such functions exist), we obtain other solutions. The way to select the right solution is to carefully manage the boundary.

In the case of a free boundary, the variational calculus tells us that minimizing $\mathcal{F}_{L_2}(z)$ requires that \eqref{eq:24} is imposed on the boundary, but the solution is still non-unique, since it is known up to an additive constant. The same conclusion holds true for any Neumann boundary condition. On the other hand, as soon as $\Omega$ is bounded, a Dirichlet boundary condition ensures existence and uniqueness of the solution\footnote{See for instance Theorem 2.4.2.6 on page 125 in \cite{Grisvard:1985} for the Dirichlet case, and Theorem 2.4.2.7 on page 126 in \cite{Grisvard:1985} for the Neumann case.}.


\section{Two Pioneering Normal Integration Methods} \label{sec:3}

Before a more exhaustive review, we first make a thorough study of two pioneering normal integration methods which have very different peculiarities. This will allow us to detect the most important properties that one may expect from any method of integration.


\subsection{Horn and Brooks' Method} \label{sec:3.1}

A well-known method due to Horn and Brooks \cite{Horn:1986a}, which we denote by $\mathcal{M}_{\text{HB}}$, attempts to solve the following discrete analogue of the Poisson equation \eqref{eq:23}, where the $(u,v)$ denote the \emph{pixels} of a square 2D-grid:
\begin{equation}
	\begin{array}{l}
		z_{u+1,v}+z_{u-1,v}+z_{u,v+1}+z_{u,v-1} - 4\,z_{u,v} \\
		\qquad\qquad\qquad = \frac{p_{u+1,v}-p_{u-1,v}}{2} + \frac{q_{u,v+1}-q_{u,v-1}}{2}
	\end{array}
\label{eq:31}
\end{equation}
The left-hand and right-hand sides of \eqref{eq:31} are second order finite differences approximations of the Laplacian and of the divergence, respectively. As stated in \cite{Harker:2015a}, other approximations can be considered, as long as the orders of the finite differences are consistent.

In \cite{Horn:1986a}, the equations \eqref{eq:31} are solved using a Jacobi iteration, for $(u,v) \in \mathring{\Omega}$ i.e., for the pixels $(u,v) \in \Omega$ whose four nearest neighbors are inside $\Omega$:
\begin{equation}
	\begin{array}{l}
		z_{u,v}^{(k+1)} = \frac{z_{u+1,v}^{(k)}+z_{u-1,v}^{(k)}+z_{u,v+1}^{(k)}+z_{u,v-1}^{(k)}}{4} \\
		\qquad\qquad\qquad -\frac{p_{u+1,v}-p_{u-1,v}}{8}-\frac{q_{u,v+1}-q_{u,v-1}}{8}
	\end{array}
\label{eq:32}
\end{equation}
The values of $z$ for the pixels $(u,v) \in \partial\Omega$ can be deduced from a discrete analogue of the natural boundary condition~\eqref{eq:24}, ``provided that the boundary curve is polygonal, with horizontal and vertical segments only''.

It is standard to show the convergence of this scheme \cite{Analyse_num}, whatever the initialization, but it converges very slowly if the initialization is far from the solution \cite{Durou:2007a}.

However, it is not so easy to discretize the natural boundary condition properly, because many cases have to be considered (see, for instance, \cite{Breuss:2017}).


\subsection{Improvement of Horn and Brooks' Method} \label{sec:3.2}

Horn and Brooks' method can be extended in order to more properly manage the natural boundary condition, by discretizing the functional $\mathcal{F}_{L_2}(z)$ defined in \eqref{eq:21}, and then solving the optimality condition, rather than discretizing the continuous optimality condition \eqref{eq:23}.

This simple idea allowed Durou and Courteille to design in \cite{Durou:2007a} an improved version of $\mathcal{M}_{\text{HB}}$, denoted by $\mathcal{M}_{\text{DC}}$, which attempts to minimize the following discrete approximation of $F_{L_2}(z)$:
\begin{equation}
	\begin{array}{l}
		F_{L_2}(\*z) = \sum_{(u,v) \in \Omega_1}\left[\frac{z_{u+1,v}-z_{u,v}}{\delta}-\frac{p_{u+1,v}+p_{u,v}}{2}\right]^2 \\
		\qquad\quad\,\, + \sum_{(u,v) \in \Omega_2}\left[\frac{z_{u,v+1}-z_{u,v}}{\delta}-\frac{q_{u,v+1}+q_{u,v}}{2}\right]^2
	\end{array}
\label{eq:30}
\end{equation}
where $\delta$ is the distance between neighboring pixels (from now on, the scale is chosen so that $\delta = 1$), $\Omega_1$ and $\Omega_2$ contain the pixels $(u,v) \in \Omega$ such that $(u+1,v)\in\Omega$ or $(u,v+1)\in\Omega$, respectively, and $\*z = [z_{u,v}]_{(u,v) \in \Omega}$.

In \eqref{eq:30}, the values of $p$ and $q$ are averaged using forward finite differences, in order to ensure the equivalence with $\mathcal{M}_{\text{HB}}$. Indeed, one gets from \eqref{eq:30} and from the characterization $\nabla F_{L_2} = 0$ of an extremum, the same optimality condition \eqref{eq:31} as discretized by Horn and Brooks, for the pixels $(u,v) \in \mathring{\Omega}$.

However, handling the boundary is much simpler than with $\mathcal{M}_{\text{HB}}$, because the appropriate discretization along the boundary naturally arises from the optimality condition $\nabla F_{L_2} = 0$. Indeed, for a pixel $(u,v) \in \partial \Omega$, the equation $\partial F_{L_2}/\partial z_{u,v} = 0$ does not take the form \eqref{eq:31} any more. Figure \ref{fig:4} shows the example of a pixel $(u,v) \in \partial\Omega$ such that $(u+1,v)$ and $(u,v+1)$ are inside~$\Omega$, while $(u-1,v)$ and $(u,v-1)$ are outside $\Omega$.
\begin{figure}[htb]
\begin{center}
	\scalebox{0.6}{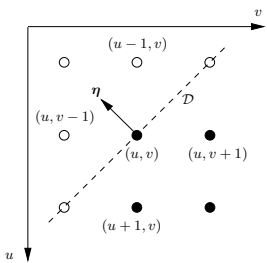}
\end{center}
\caption{Only the black pixels are inside $\Omega$. The straight line $\mathcal{D}$ is a plausible approximation of the tangent to $\partial\Omega$ at $(u,v)$.}
\label{fig:4}
\end{figure}

In this case, Equation \eqref{eq:31} must be replaced with:
\begin{equation}
	\begin{array}{l}
		z_{u+1,v}+z_{u,v+1} - 2\,z_{u,v} \\
		\qquad\qquad\qquad = \small{\frac{p_{u+1,v}+p_{u,v}}{2}} + \frac{q_{u,v+1}+q_{u,v}}{2}
	\end{array}
\label{eq:33}
\end{equation}
Since $\boldsymbol{\eta} = -\sqrt{2}/2\,[1,1]^\top$ is a plausible unit-length normal to the boundary $\partial\Omega$ in this case, the natural boundary condition \eqref{eq:24} reads:
\begin{equation}
	\partial_u z-p+\partial_v z-q = 0
\label{eq:34}
\end{equation}
It is obvious that \eqref{eq:33} is a discrete approximation of \eqref{eq:34}. More generally, it is easily shown that the equation $\partial F_{L_2}/\partial z_{u,v} = 0$, for any $(u,v) \in \partial\Omega$, is a discrete approximation of the natural boundary condition \eqref{eq:24}.}

\begin{figure*}[htb]
\begin{center}
	\begin{tabular}{cc}
		\includegraphics[scale=0.53]{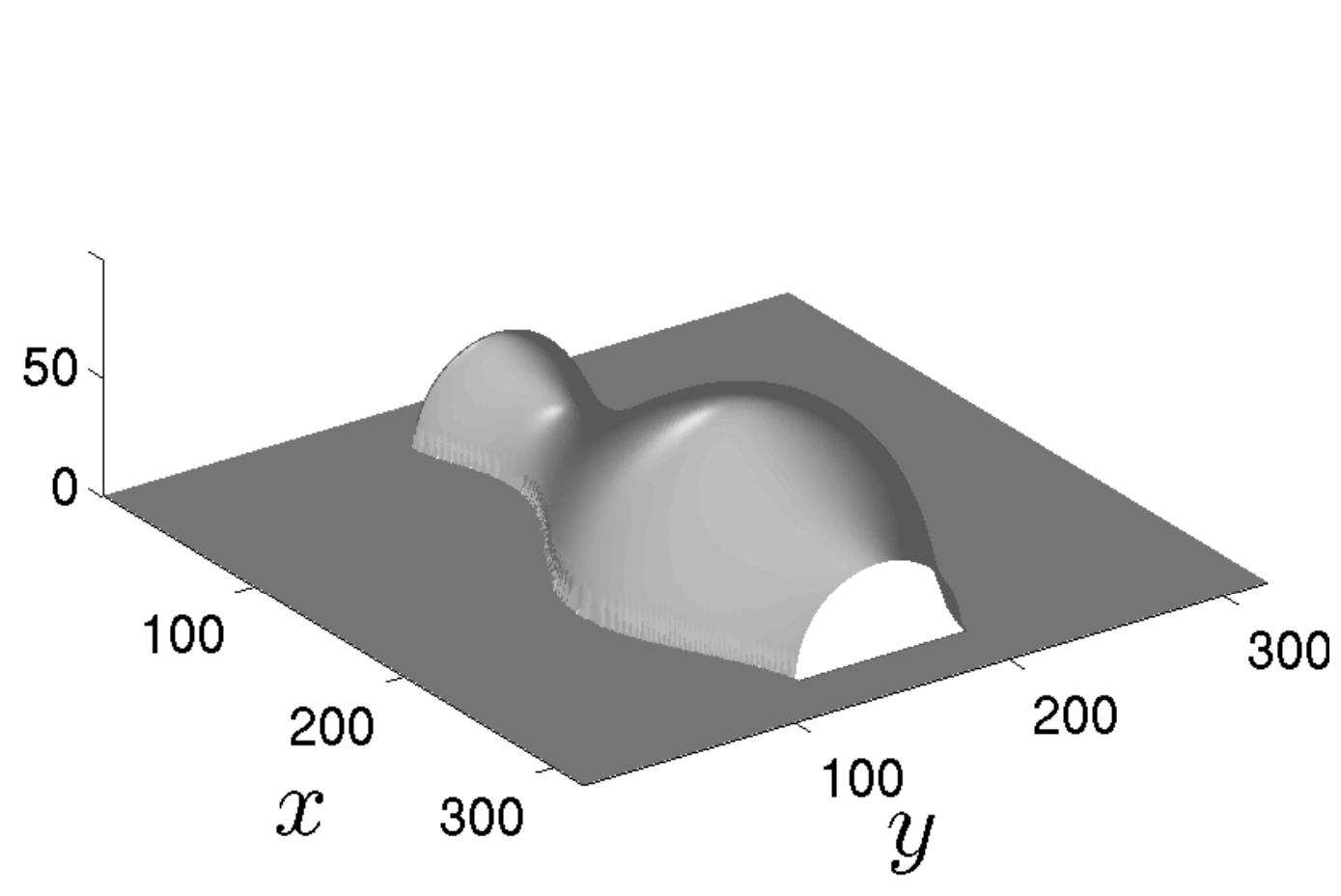} ~ & \rotatebox{90}{\includegraphics[scale=0.53]{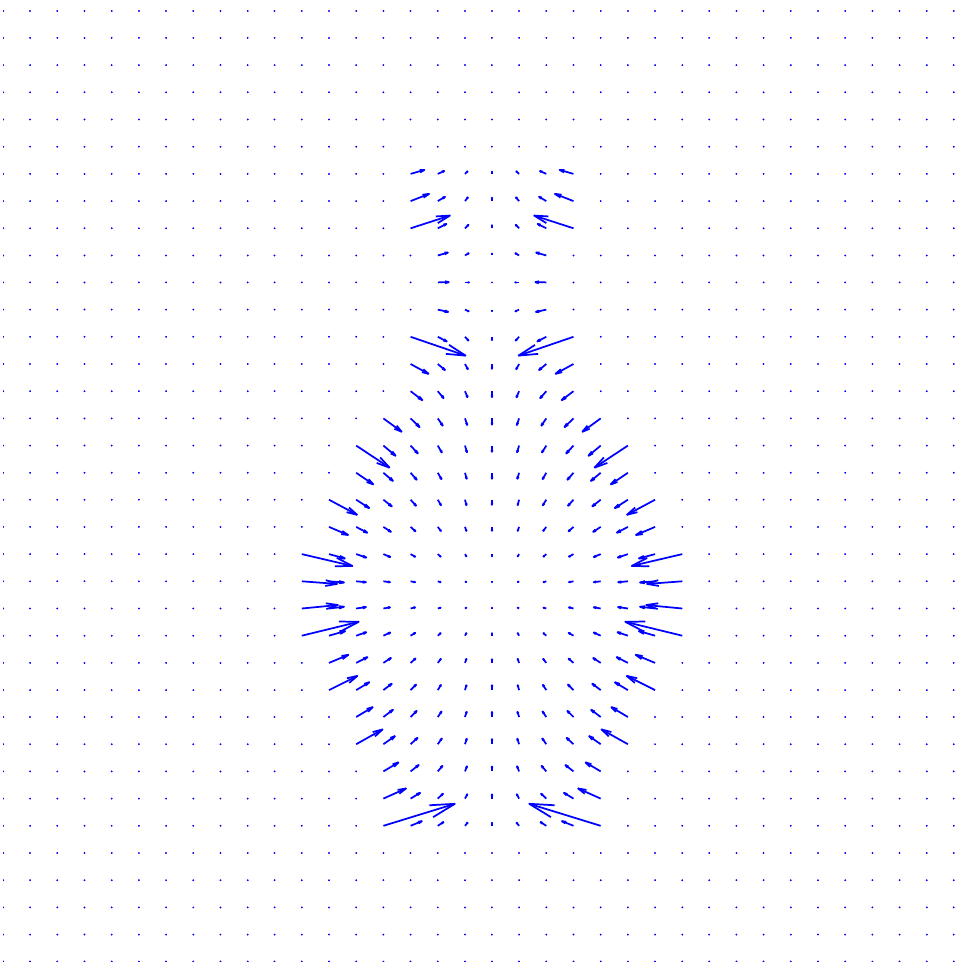}} \\
		(a) Surface $\mathcal{S}_{\text{vase}}$ ~ & (b) Gradient field $\*g_{\text{vase}}$ \\
		\includegraphics[scale=0.53]{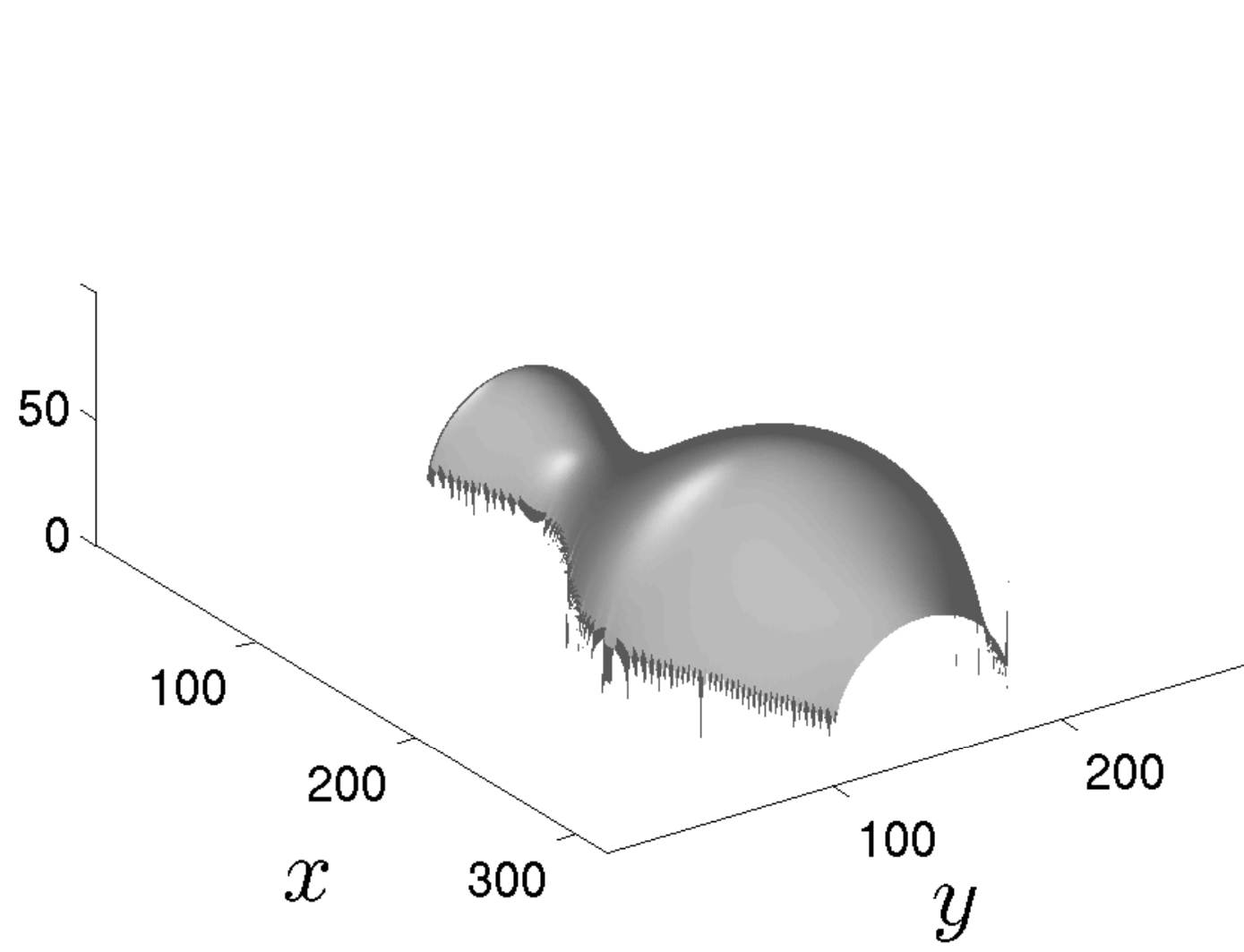} ~ & \qquad \includegraphics[scale=0.53]{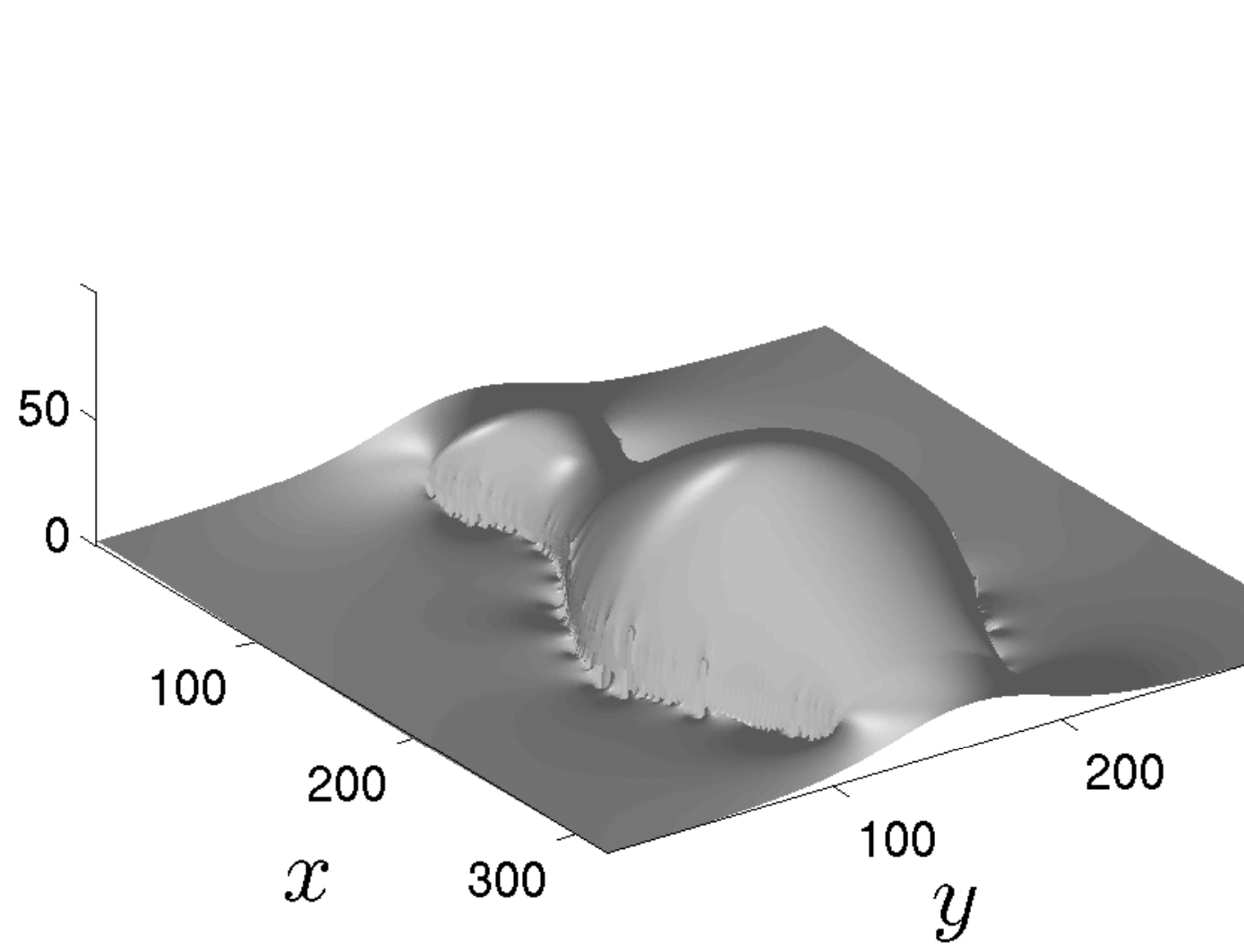} \\
		(c) $\text{RMSE} = 0.93$ ~ & (d) $\text{RMSE} = 4.51$ \\
	\end{tabular}
\end{center}
\caption{(a) Test surface $\mathcal{S}_{\text{vase}}$. (b) Gradient field $\*g_{\text{vase}}$, obtained by sampling of the analytically computed gradient of $\mathcal{S}_{\text{vase}}$. (c)~Reconstructed surface obtained by integration of $\*g_{\text{vase}}$, using as reconstruction domain the image of the vase, at convergence of $\mathcal{M}_{\text{DC}}$. (d) Same test, but on the whole grid. In both tests, the constant of integration is chosen so as to minimize the RMSE.}
\label{fig:5}
\end{figure*}

Let us test $\mathcal{M}_{\text{DC}}$ on the surface $\mathcal{S}_{\text{vase}}$ shown in Figure \ref{fig:5}-a, which models a half-vase lying on a flat ground. To this end, we sample the \emph{analytically computed} gradient of $\mathcal{S}_{\text{vase}}$ on a regular grid of size $312\times312$, which provides the gradient field $\*g_{\text{vase}}$ (see Figure \ref{fig:5}-b). We suppose that a preliminary segmentation allows us to use as reconstruction domain the image of the vase, which constitutes an additional datum. The reconstructed surface obtained at convergence of $\mathcal{M}_{\text{DC}}$ is shown in Figure \ref{fig:5}-c.

On the other hand, it is well-known that quadratic regularization is not well-adapted to discontinuities. Let us now test $\mathcal{M}_{\text{DC}}$ using as reconstruction domain the whole grid of size $312\times312$, which contains discontinuities at the top and at the bottom of the vase. The reconstructed surface at convergence is shown in Figure \ref{fig:5}-d. It is not satisfactory, since the discontinuities are not preserved. This is numerically confirmed by the RMSE, which is much higher than that of Figure \ref{fig:5}-c.

We know that removing the flat ground from the reconstruction domain suffices to reach a better result (see Figure \ref{fig:5}-c), since this eliminates any depth discontinuity. However, this requires a preliminary segmentation, which is known to be a hard task, and also requires that the integration can be carried out on a \emph{non-rectangular reconstruction domain}.

Otherwise, it is necessary to appropriately handle the depth discontinuities, in order to limit the bias. The lack of integrability of the vector field $[p,q]^\top$ is a basic idea to detect the discontinuities, as shown in our companion paper \cite{Queau:2017}. Unfortunately, this characterization of the discontinuities is neither necessary nor sufficient. On the one hand, shadows can induce outliers if $[p,q]^\top$ is estimated via photometric stereo \cite{Reddy:2009a}, which can therefore be non-integrable along shadow limits, even in the absence of depth discontinuities. On the other hand, the vector field $[p,q]^\top$ of a scale seen from above is uniform, i.e. perfectly integrable.


\subsection{Frankot and Chellappa's Method} \label{sec:3.3}

A more general approach to overcome the possible non-integrability of the gradient field $\*g = [p,q]^\top$ is to first define a set $\mathcal{I}$ of integrable vector fields i.e., of vector fields of the form $\nabla z$, and then compute the projection $\nabla \bar{z}$ of $\*g$ on $\mathcal{I}$ i.e., the vector field $\nabla \bar{z}$ of $\mathcal{I}$ the closest to $\*g$, according to some norm. Afterwards, the (approximate) solution of Equation \eqref{eq:20} is easily obtained using \eqref{eq:5}, \eqref{eq:9} or \eqref{eq:18}, since $\nabla \bar{z}$ is integrable.

Nevertheless, the boundary conditions can be complicated to manage, because $\mathcal{I}$ depends on which boundary condition is imposed (including the case of the natural boundary condition). It is noticed in \cite{Agrawal:2006a} that minimizing the functional $\mathcal{F}_{L_2}(z)$ amounts to following this general approach, in the case where $\mathcal{I}$ contains all integrable vector fields and the Euclidean norm is used.

The most cited normal integration method, due to Frankot and Chellappa \cite{Frankot:1988a}, follows this approach in the case where the Fourier basis is considered. Let us use the standard definition of the Fourier transform:
\begin{equation}
	\hat{f}(\omega_u,\omega_v) = \displaystyle\iint\limits_{(u,v)\in\mathbb{R}^2} f(u,v) \, e^{-\mathsf{j} \, \omega_u\,u} \, e^{-\mathsf{j} \, \omega_v\,v}\, \mathrm{d}u \,\mathrm{d}v
\label{eq:35}
\end{equation}
where $(\omega_u,\omega_v) \in \mathbb{R}^2$. Computing the Fourier transforms of both sides of \eqref{eq:23}, we obtain:
\begin{equation}
	\small{-(\omega_u^2+\omega_v^2)\, \hat{z}(\omega_u,\omega_v) = \mathsf{j}\,\omega_u\,\hat{p}(\omega_u,\omega_v)+\mathsf{j}\,\omega_v\,\hat{q}(\omega_u,\omega_v)}
\label{eq:36}
\end{equation}
In Equation \eqref{eq:36}, the data $\hat{p}$ and $\hat{q}$, as well as the unknown $\hat{z}$, depend on the variables $\omega_u$ and $\omega_v$. For any $(\omega_u,\omega_v)$ such that $\omega_u^2+\omega_v^2\neq 0$, this equation gives us the following expression of $\hat{z}(\omega_u,\omega_v)$:
\begin{equation}
	\hat{z}(\omega_u,\omega_v) = \frac{\omega_u\,\hat{p}(\omega_u,\omega_v)+\omega_v\,\hat{q}(\omega_u,\omega_v)}{\mathsf{j}\,(\omega_u^2+\omega_v^2)}
\label{eq:37}
\end{equation}
Indeed, computing the inverse Fourier transform of \eqref{eq:37} will provide us with a solution of~\eqref{eq:23}. This method of integration \cite{Frankot:1988a}, which we denote by $\mathcal{M}_{\text{FC}}$, is very fast thanks to the FFT algorithm.

The definition of the Fourier transform may be confusing, because several definitions exist. Instead of pulsations $(\omega_u,\omega_v)$, frequencies $(n_u,n_v)$ can be used. Knowing that $(\omega_u,\omega_v) = 2\pi (n_u,n_v)$, Equation \eqref{eq:37} then becomes: 
\begin{equation}
	\hat{z}(n_u,n_v) = \frac{n_u\,\hat{p}(n_u,n_v)+n_v\,\hat{q}(n_u,n_v)}{2\pi\,\mathsf{j}\,(n_u^2+n_v^2)}
\label{eq:37bis}
\end{equation}
It is written in \cite{Ng_HS:2007a} that the accuracy of Frankot and Chellappa's method ``relies on a good input scale''. In fact, it only happens that, in a publicly available code of this method, the $2\pi$ coefficient in \eqref{eq:37bis} is missing.

Since the right-hand side of Equation \eqref{eq:37} is not defined if $(\omega_u,\omega_v) = (0,0)$, Frankot and Chellappa assert that it ``simply means that we cannot recover the average value of $z$ without some additional information''. This is true but incomplete, because $\mathcal{M}_{\text{FC}}$ provides the solution of Equation \eqref{eq:23} \emph{up to the addition of a harmonic function} over $\Omega$. Saracchini et al. note in \cite{Saracchini:2012a} that ``the homogeneous version of [\eqref{eq:23}] is satisfied by an arbitrary linear function of position $z(u,v) = a\,u+b\,v+c$, which when added to any solution of [\eqref{eq:23}] will yield infinitely many additional solutions''. Affine functions are harmonic indeed, but we know from Section \ref{sec:2.3} that many other harmonic functions also exist.

Applying $\mathcal{M}_{\text{FC}}$ to $\*g_{\text{vase}}$ would give the same result as that of Figure \ref{fig:5}-d, but much faster, when the reconstruction domain is equal to the whole grid. On the other hand, such a result as that of Figure \ref{fig:5}-c could not be reached, since $\mathcal{M}_{\text{FC}}$ is not designed to manage a non-rectangular domain $\Omega$.

\begin{figure*}[!ht]
\begin{center}
	\begin{tabular}{cc}
		\includegraphics[width=0.48 \linewidth,height=0.2\linewidth]{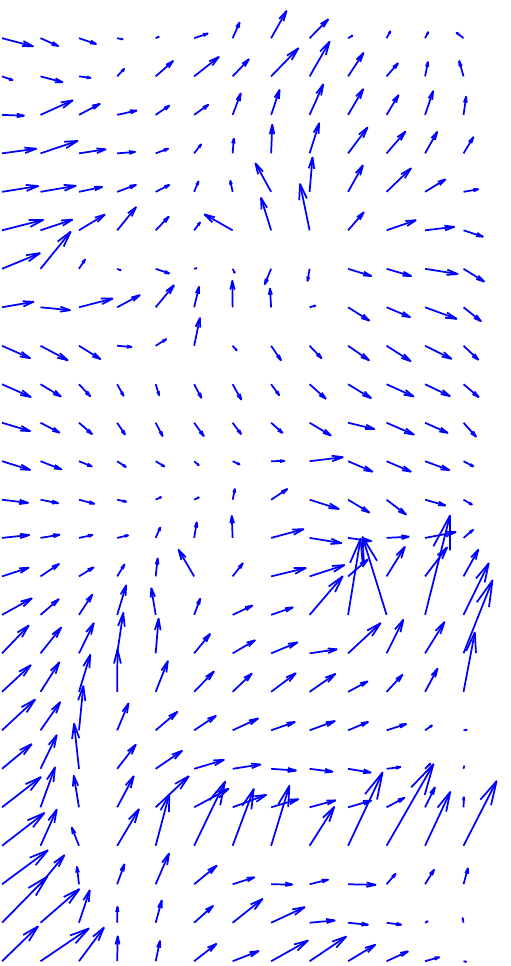} &
		\includegraphics[width=0.48 \linewidth,trim=0 0 0 1cm, clip]{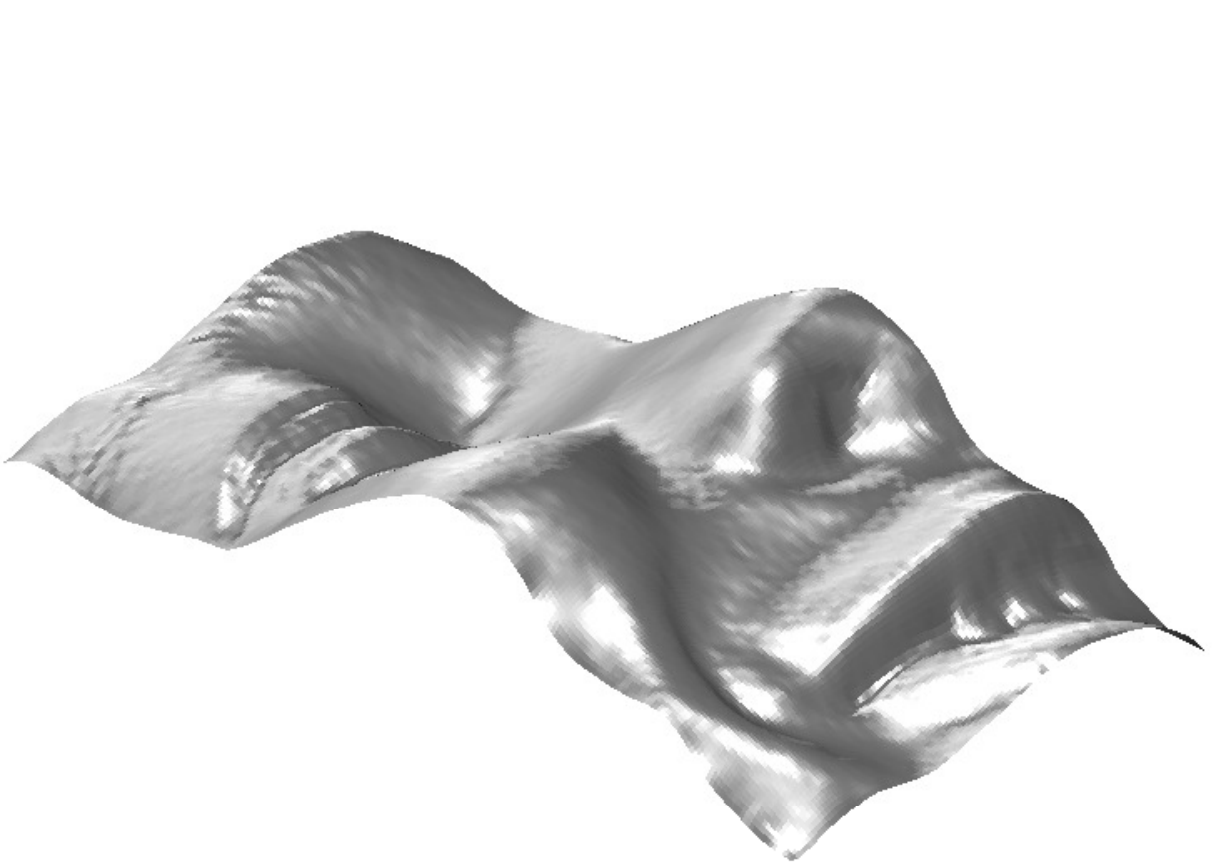} \\
		(a) & (b) \\
	\end{tabular}
\end{center}
\caption{(a) Gradient field $\*g_{\text{face}}$ of a face estimated via photometric stereo. (b) Reconstructed surface obtained by integration of $\*g_{\text{face}}$, using Frankot and Chellappa's method \cite{Frankot:1988a}. Since a periodic boundary condition is assumed, the depth is forced to be the same on the cheek and on the nose. As a consequence, this much distorts the result.}
\label{fig:6}
\end{figure*}

Besides, $\mathcal{M}_{\text{FC}}$ works well if and only if the surface to be reconstructed is periodic. This clearly appears in the example of Figure \ref{fig:6}: the gradient field $\*g_{\text{face}}$ of a face (see Figure \ref{fig:6}-a), estimated via photometric stereo \cite{Woodham:1980a}, is integrated using $\mathcal{M}_{\text{FC}}$, which results in the reconstructed surface shown in Figure \ref{fig:6}-b. Since the face is non-periodic, but the depth is forced to be the same on the left and right edges of the face i.e., on the cheek and on the nose, the result is much distorted.

This failure of $\mathcal{M}_{\text{FC}}$ was first exhibited by Harker and O'Leary in \cite{Harker:2008a}, who show on some examples that the solution provided by $\mathcal{M}_{\text{FC}}$ is not always a minimizer of the functional $\mathcal{F}_{L_2}(z)$. They moreover explain: ``The fact that the solution [of Frankot and Chellappa] is constrained to be periodic leads to a systematic bias in the solution'' (this periodicity is clearly visible on the recovered surface of Figure \ref{fig:6}-b). Harker and O'Leary also conclude that ``any approach based on the Euler-Lagrange equation is only valid for a few special cases''. The improvements of $\mathcal{M}_{\text{FC}}$ that we describe in the next section make this assertion highly questionable.

\subsection{Improvements of Frankot and Chellappa's Method} \label{sec:3.4}

A first improvement of $\mathcal{M}_{\text{FC}}$ suggested by Simchony, Chellappa and Shao in \cite{Simchony:1990a} amounts to solving the discrete approximation \eqref{eq:31} of the Poisson equation using the discrete Fourier transform, instead of discretizing the solution \eqref{eq:37} of the Poisson equation \eqref{eq:23}.

Consider a rectangular 2D domain $\Omega = [0,d_u]\times[0,d_v]$, and choose a lattice of equally spaced points $(u\frac{d_u}{m},v\frac{d_v}{n})$, $u\in[0,m]$, $v\in[0,n]$. Let us denote by $f_{u,v}$ the value of a function $f : \Omega \rightarrow \mathbb{R}$ at $(u\frac{d_u}{m},v\frac{d_v}{n})$. The standard definition of the \emph{discrete} Fourier transform of $f$ is as follows, for $k\in[0,m-1]$ and $l\in[0,n-1]$:
\begin{equation}
	\hat{f}_{k,l} = \sum_{u=0}^{m-1} \sum_{v=0}^{n-1} f_{u,v} \, \displaystyle e^{-\mathsf{j} 2\pi\frac{uk}{m}}\, \displaystyle e^{-\mathsf{j} 2\pi\frac{vl}{n}}
\label{eq:38}
\end{equation}
The inverse transform of \eqref{eq:38} reads:
\begin{equation}
	f_{u,v} = \frac{1}{mn} \, \sum_{k=0}^{m-1} \sum_{l=0}^{n-1} \hat{f}_{k,l} \, \displaystyle e^{+\mathsf{j} 2\pi\frac{ku}{m}}\, \displaystyle e^{+\mathsf{j} 2\pi\frac{lv}{n}}
\label{eq:39}
\end{equation}

Replacing any term in \eqref{eq:31} by its inverse discrete Fourier transform of the form \eqref{eq:39}, and knowing that the Fourier family is a basis, we obtain:
\begin{equation}
	\begin{array}{ll}
		2 \left[\cos\left(2\pi \frac{k}{m}\right) + \cos\left(2\pi \frac{l}{n}\right) - 2 \right] \hat{z}_{k,l} \\
		\qquad\qquad = \mathsf{j} \left[\sin\left(2\pi \frac{k}{m}\right) \, \hat{p}_{k,l} + \sin\left(2\pi \frac{l}{n}\right) \, \hat{q}_{k,l} \right]
	\end{array}
\label{eq:40}
\end{equation}
The expression in brackets of the left-hand side is zero if and only if $(k,l) = (0,0)$. As soon as $(k,l)\neq(0,0)$, Equation \eqref{eq:40} provides us with the expression of $\hat{z}_{k,l}$:
\begin{equation}
	\hat{z}_{k,l} = \frac{\sin\left(2\pi \frac{k}{m}\right) \, \hat{p}_{k,l} + \sin\left(2\pi \frac{l}{n}\right) \, \hat{q}_{k,l}}
					{4 \mathsf{j} \left[\sin^2\left(\pi \frac{k}{m}\right) + \sin^2\left(\pi \frac{l}{n}\right)\right]}
\label{eq:41}
\end{equation}
which is rewritten by Simchony et al. as follows:
\begin{equation}
	\hat{z}_{k,l} = \frac{\sin\left(2\pi \frac{k}{m}\right) \, \hat{p}_{k,l} + \sin\left(2\pi \frac{l}{n}\right) \, \hat{q}_{k,l}}
					{\mathsf{j} \left[\frac{\sin^2\left(2\pi \frac{k}{m}\right)}{\cos^2(\pi \frac{k}{m})} + \frac{\sin^2\left(2\pi \frac{l}{n}\right)}{\cos^2(\pi \frac{l}{n})}\right]}
\label{eq:42}
\end{equation}

Comparing \eqref{eq:35} and \eqref{eq:38} shows us that $\omega_u$ corresponds to $2\pi \frac{k}{m}$ and $\omega_v$ to $2\pi \frac{l}{n}$. Using these correspondences, we would expect to be able to identify \eqref{eq:37} and \eqref{eq:42}. This is true if $\sin\left(2\pi \frac{k}{m}\right)$ and $\sin\left(2\pi \frac{l}{n}\right)$ tend toward 0, and $\cos^2\left(\pi \frac{k}{m}\right)$ and $\cos^2\left(\pi \frac{l}{n}\right)$ tend toward 1, which occurs if $k$ takes either the first values or the last values inside $[0,m-1]$, and the same for $l$ inside $[0,n-1]$, which is interpreted by Simchony et al. as follows: ``At low frequencies our result is similar to the result obtained in \cite{Frankot:1988a}. At high frequencies we attenuate the corresponding coefficients since our discrete operator has a low-pass filter response [...] the surface $z$ obtained in \cite{Frankot:1988a} may suffer from high frequency oscillations''. Actually, the low values of $k$ and $l$ correspond to the lowest values of $\omega_u$ and $\omega_v$ inside $\mathbb{R}^+$, and the high values of $k$ and $l$ may be interpreted as the lowest absolute values of $\omega_u$ and $\omega_v$ inside $\mathbb{R}^-$.

Moreover, using the inverse discrete Fourier transform \eqref{eq:39}, the solution of \eqref{eq:31} which follows from \eqref{eq:41} will be periodic in $u$ with period $d_u$, and in $v$ with period $d_v$. This means that $z_{m,v} = z_{0,v}$, for $v\in[0,n]$, and $z_{u,n} = z_{u,0}$, for $u\in[0,m]$. The discrete Fourier transform is therefore appropriate only for problems which satisfy periodic boundary conditions.

Much more relevant improvements of $\mathcal{M}_{\text{FC}}$ due to Simchony et al. are to suggest that using the discrete sine transform or the discrete cosine transform is appropriate if the problem involves, respectively, Dirichlet or Neumann boundary conditions.

\paragraph{Dirichlet Boundary Condition}

It is easily checked that any function that is expressed as an inverse discrete sine transform of the following form\footnote{Whereas the Fourier transform coefficients of $f$ are denoted by $\hat{f}$, the sine and cosine transform coefficients are denoted by $\bar{f}$ and $\bar{\bar{f}}$, respectively.}:
\begin{equation}
	f_{u,v} = \frac{4}{mn}\, \sum_{k=1}^{m-1} \sum_{l=1}^{n-1} \bar{f}_{k,l} \, \sin\left(\pi\frac{ku}{m}\right)\, \sin\left(\pi\frac{lv}{n}\right)
\label{eq:43}
\end{equation}
satisfies the homogeneous Dirichlet condition $f_{u,v} = 0$ on the boundary of the discrete domain $\Omega = [0,m] \times [0,n]$ i.e., for $u = 0$, $u = m$, $v = 0$, and $v = n$.

To solve \eqref{eq:31} on a rectangular domain $\Omega$ with the homogeneous Dirichlet condition $z_{u,v} = 0$ on $\partial\Omega$, we can therefore write $z_{u,v}$ as in \eqref{eq:43}. This is still possible for a non-homogeneous Dirichlet boundary condition:
\begin{equation}
	z_{u,v} = b^D_{u,v} \quad \text{for } (u,v) \in \partial\Omega
\label{eq:44}
\end{equation}

A first solution would be to solve a pair of problems. We could search for both a solution $z^0_{u,v}$ of the equations \eqref{eq:31} satisfying the homogeneous Dirichlet boundary condition, thus taking the form \eqref{eq:43}, and a harmonic function $h_{u,v}$ on $\Omega$ satisfying the boundary condition \eqref{eq:44}. Then, $z^0_{u,v}+h_{u,v}$ is a solution of the equations \eqref{eq:31} satisfying this boundary condition.

But it is much easier to replace $z_{u,v}$ with $z'_{u,v}$, such that $z'_{u,v} = z_{u,v}$ everywhere on ${\Omega}$, except on its boundary $\partial \Omega$ where $z'_{u,v} = z_{u,v}-b^D_{u,v}$. The Dirichlet boundary condition satisfied by $z'_{u,v}$ is homogeneous, so we can actually write $z'_{u,v}$ under the form \eqref{eq:43}.

In practice, we just have to change the right-hand side $g_{u,v}$ of Equation \eqref{eq:31} for the pixels $(u,v) \in \Omega$ which are adjacent to $\partial\Omega$. Either one or two neighbors of these pixels lie on $\partial \Omega$. For example, only the neighbor $(0,v)$ of pixel $(1,v)$ lies on $\partial \Omega$, for $v \in [2,n-2]$. In such a pixel, the right-hand side of Equation \eqref{eq:31} must be replaced with:
\begin{equation}
	g^D_{1,v} = \dfrac{p_{2,v}-p_{0,v}}{2} + \dfrac{q_{1,v+1}-q_{1,v-1}}{2} - b^D_{0,v}
\label{eq:45}
\end{equation}
On the other hand, amongst the four neighbors of the corner pixel $(1,1)$, both $(0,1)$ and $(1,0)$ lie on $\partial \Omega$. Therefore, the right-hand side of Equation \eqref{eq:31} must be modified as follows, in this pixel:
\begin{equation}
	g^D_{1,1} = \dfrac{p_{2,1}-p_{0,1}}{2} + \dfrac{q_{1,2}-q_{1,0}}{2} - b^D_{0,1} - b^D_{1,0}
\end{equation}

Knowing that the products of sine functions in \eqref{eq:43} form a linearly independent family, we get from \eqref{eq:31} and \eqref{eq:43}, $\forall (k,l) \in [1,m-1] \times [1,n-1]$:
\begin{equation}
	\bar{z}'_{k,l} = - \frac{\bar{g}^D_{k,l}}{4 \, \left(\sin^2\frac{\pi k}{2m}+\sin^2\frac{\pi l}{2n}\right)},
\label{eq:46}
\end{equation}
From \eqref{eq:46}, we easily deduce $z'_{u,v}$ using the inverse discrete sine transform \eqref{eq:43}, thus $z_{u,v}$.

\paragraph{Neumann Boundary Condition}

The reasoning is similar, yet a little bit trickier, in the case of a Neumann boundary condition. Any function that is expressed as an inverse discrete cosine transform of the following form:
\begin{equation}
	f_{u,v} = \frac{4}{mn}\, \sum_{k=0}^{m-1} \sum_{l=0}^{n-1} \bar{\bar{f}}_{k,l} \, \cos\left(\pi\frac{ku}{m}\right)\, \cos\left(\pi\frac{lv}{n}\right)
\label{eq:47}
\end{equation}
satisfies the homogeneous Neumann boundary condition $\nabla f_{u,v} \cdot \boldsymbol{\eta}_{u,v} = 0$ on $\partial\Omega$, where $\boldsymbol{\eta}_{u,v}$ is the outer unit-length normal to $\partial \Omega$ in pixel $(u,v)$.
%
To solve \eqref{eq:31} on a rectangular domain $\Omega$ with the homogeneous Neumann condition on $\partial\Omega$, we can thus write $z_{u,v}$ as in \eqref{eq:47}. Consider now a non-homogeneous Neumann boundary condition:
\begin{equation}
	\nabla z_{u,v} \cdot \boldsymbol{\eta}_{u,v} = b^N_{u,v} \quad \text{for } (u,v) \in \partial\Omega
\label{eq:49}
\end{equation}

A similar trick as before consists in defining an auxiliary function $z''_{u,v}$ such that $z''_{u,v}$ is equal to $z_{u,v}$ on $\Omega$, but differs from $z_{u,v}$ outside $\Omega$ and satisfies the homogeneous Neumann boundary condition on $\partial \Omega$. Let us first take the example of a pixel $(0,v) \in \partial \Omega$, $v \in [1,n-1]$. By discretizing~\eqref{eq:49} using first-order finite differences, it is easily verified that $z''$ satisfies the homogeneous Neumann boundary condition in such a pixel if $z''_{-1,v} = z_{-1,v}+b^N_{0,v}$. Similar definitions of $z''$ arise on the other three edges of ${\Omega}$.

Let us now take the example of a corner of $\Omega$, for instance pixel $(0,0)$. Knowing that $\boldsymbol{\eta}_{0,0} = -\frac{1}{\sqrt{2}}\,[1,1]^\top$, one can easily check, by discretizing~\eqref{eq:49} using first-order finite differences, that appropriate modifications of $z$ in this pixel are $z''_{-1,0} = z_{-1,0} + \frac{\sqrt{2}}{2} \, b^N_{0,0}$ and $z''_{0,-1} = z_{0,-1} + \frac{\sqrt{2}}{2} \, b^N_{0,0}$. With these modifications, the function $z''_{u,v}$ indeed satisfies the homogeneous Neumann boundary condition in pixel $(0,0)$.

In practice, we just have to change the right-hand side $g_{u,v}$ of Equation \eqref{eq:31} for the pixels $(u,v) \in \partial\Omega$. Either one or two neighbors of these pixels lie outside $\Omega$. For example, only the neighbor $(-1,v)$ of pixel $(0,v)$ lies outide $\Omega$, for $v \in [1,n-1]$. In such a pixel, the right-hand side of Equation \eqref{eq:31} must be replaced with:
\begin{equation}
	g^N_{0,v} = \dfrac{p_{1,v}-p_{-1,v}}{2} + \dfrac{q_{0,v+1}-q_{0,v-1}}{2} - b^N_{0,v}
\label{eq:50}
\end{equation}
A problem with this expression is that $p_{-1,v}$ is unknown. Using a first-order discretization of the homogeneous Neumann boundary condition (on $p$, not on $z$) $\nabla p_{0,v} \cdot \boldsymbol{\eta}_{0,v} = 0$, we obtain the approximation $p_{-1,v} \approx p_{0,v}$, and thus \eqref{eq:50} is turned into:
\begin{equation}
	g^N_{0,v} = \dfrac{p_{1,v} - p_{0,v}}{2} + \dfrac{q_{0,v+1}-q_{0,v-1}}{2} - b^N_{0,v}
\label{eq:50bis}
\end{equation}
On the other hand, amongst the four neighbors of pixel $(0,0)$, which is a corner of $\Omega$, both $(0,-1)$ and $(-1,0)$ lie outside $\Omega$. Therefore, the right-hand side of Equation \eqref{eq:31} must be modified as follows:
\begin{equation}
	g^N_{0,0} = \dfrac{p_{1,0} - p_{0,0}}{2} + \dfrac{q_{0,1} - q_{0,0}}{2} - \sqrt{2} \, b^N_{0,0} \\
\end{equation}

Knowing that the products of cosine functions in \eqref{eq:47} form a linearly independent family, we get from \eqref{eq:31} and \eqref{eq:47}, $\forall (k,l) \in [0,m-1] \times [0,n-1]$:
\begin{equation}
	\bar{\bar{z}}''_{k,l} = - \frac{\bar{\bar{g}}^N_{k,l}}{4 \, \left(\sin^2\frac{\pi k}{2m}+\sin^2\frac{\pi l}{2n}\right)},
\label{eq:52}
\end{equation}
except for $(k,l) = (0,0)$. Indeed, we cannot determine the coefficient $\bar{\bar{z}}''_{0,0}$, which simply means that the solution of the Poisson equation using a Neumann boundary condition (as, for instance, the natural boundary condition) is computable up to an additive constant, because the term which corresponds to the coefficient $\bar{\bar{f}}_{0,0}$ in the double sum of \eqref{eq:47} does not depend on $(u,v)$. Keeping this point in mind, we can deduce $z''_{u,v}$ from \eqref{eq:52} using the inverse discrete cosine transform \eqref{eq:47}, thus $z_{u,v}$.

\begin{figure*}[htb]
\begin{center}
	\begin{tabular}{cc}
		\includegraphics[width=0.48 \linewidth,trim=0 0 0 1cm, clip]{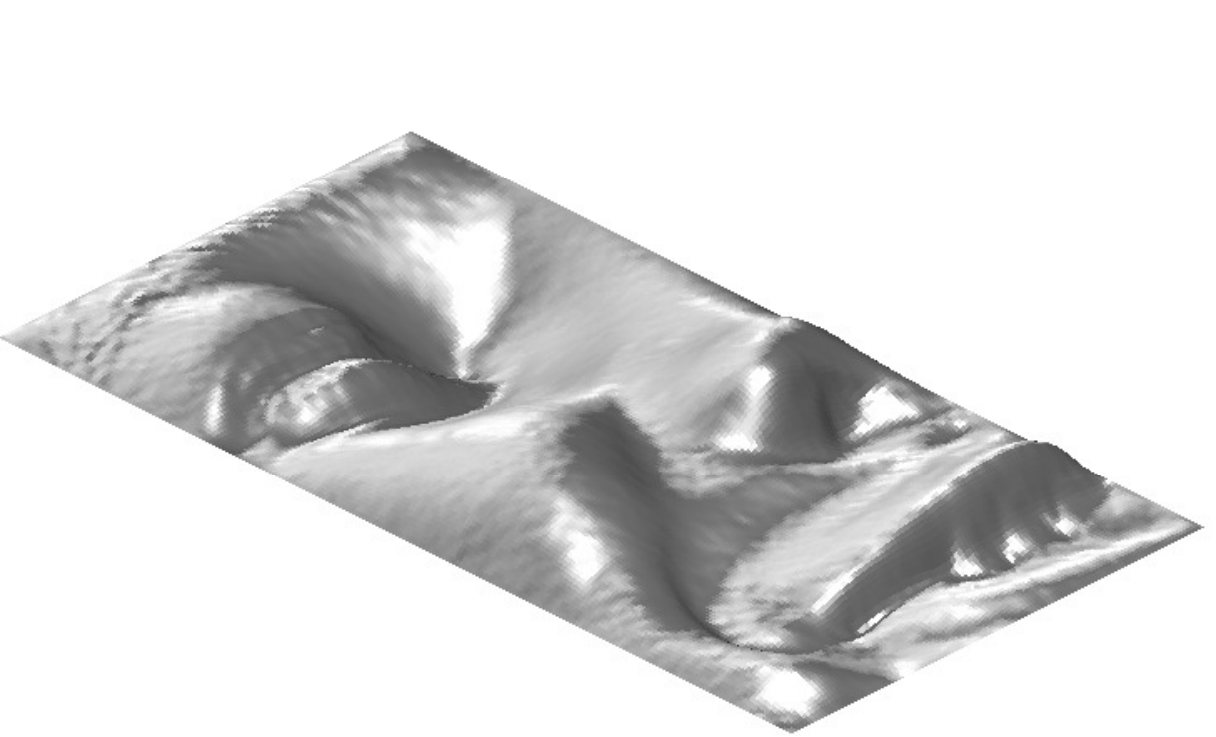} &
		\includegraphics[width=0.49 \linewidth,trim=0 0 0 1cm, clip]{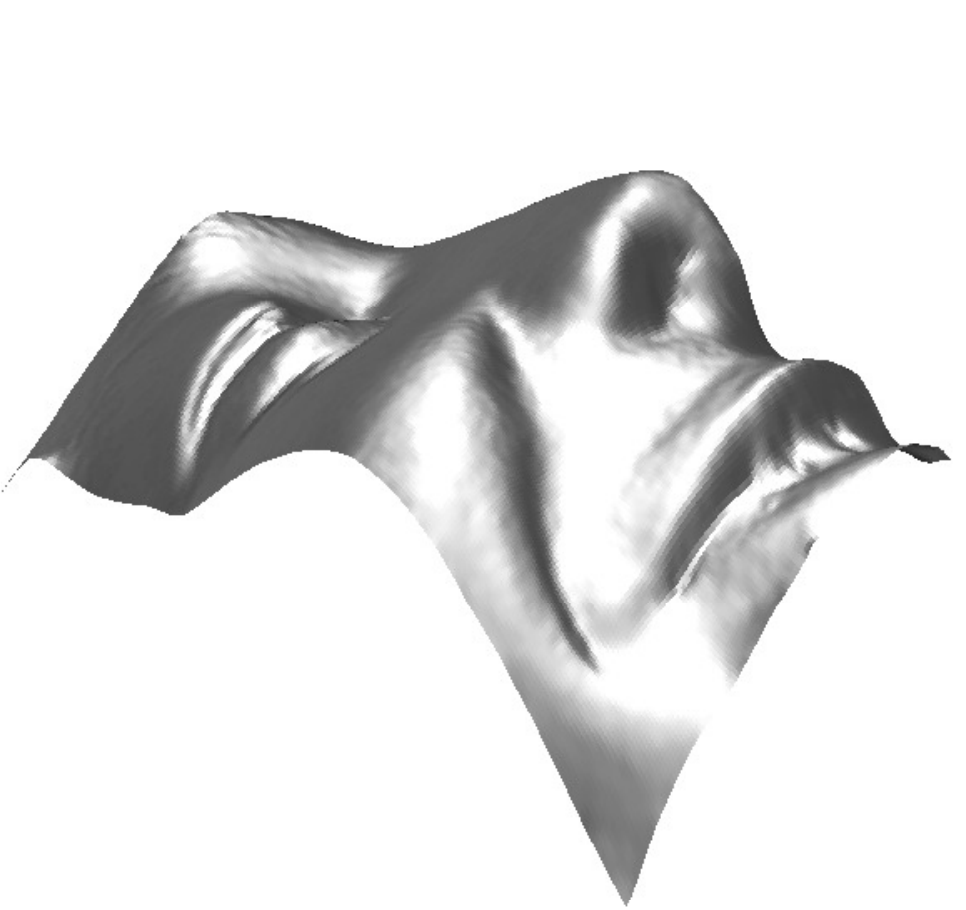} \\
		(a) & (b)
	\end{tabular}
\end{center}
\caption{Reconstructed surfaces obtained by integration of $\*g_{\text{face}}$ (see Figure \ref{fig:6}-a), using Simchony, Chellappa and Shao's method \cite{Simchony:1990a}, and imposing two different boundary conditions. (a) Since the homogeneous Dirichlet boundary condition $z = 0$ is clearly false, the surface is much distorted. (b) The (Neumann) natural boundary condition provides a much more realistic result.}
\label{fig:7}
\end{figure*}

Accordingly, the method $\mathcal{M}_{\text{SCS}}$ designed by Simchony, Chellappa and Shao in \cite{Simchony:1990a} works well even in the case of a non-periodic surface (see Figure \ref{fig:7}). Knowing moreover that this method is as fast as $\mathcal{M}_{\text{FC}}$, we conclude that it improves Frankot and Chellappa's original method a lot.

On the other hand, the useful property of the solutions \eqref{eq:43} and \eqref{eq:47} i.e., they satisfy a homogeneous Dirichlet or Neumann boundary condition, is obviously valid only if the reconstruction domain $\Omega$ is rectangular. This trick is hence not useable for any other form of domain $\Omega$. It is claimed in \cite{Simchony:1990a} that \emph{embedding techniques} can extend $\mathcal{M}_{\text{SCS}}$ to non-rectangular domains, but this is neither detailed, nor really proved. As a consequence, such a result as that of Figure \ref{fig:5}-c could not be reached applying $\mathcal{M}_{\text{SCS}}$ to the gradient field $\*g_{\text{vase}}$: the result would be the same as that of Figure \ref{fig:5}-d.


\section{{Main Normal Integration Methods}} \label{sec:4}


\subsection{{A List of Expected Properties}} \label{sec:4.1}

This may appear as a truism, but a basic requirement of 3D-reconstruction is \emph{accuracy}. Anyway, the evaluation/comparison of 3D-reconstruction methods is a difficult challenge. Firstly, it may happen that some methods require more data than the others, which makes the evaluation/compari\-son biased in some sense. Secondly, it usually happens that the choice of the benchmark has a great influence on the final ranking. Finally, it is a hard task to implement a method just from its description, which gives in practice a substantial advantage to the designers of a ranking process, whatever their methodology, in the case when they also promote their own method. In Section \ref{sec:4.3}, we will review the main existing normal integration methods. However, in accordance with these remarks, we do not intend to evaluate their accuracy. We will instead quote their main features.

In view of the detailed reviews of the methods of Horn and Brooks and of Frankot and Chellappa (see Section \ref{sec:3}), we may expect, apart from accuracy, five other properties from any normal integration method:
\begin{itemizepoint}
	\item $\mathcal{P}_{\text{Fast}}$: The desired method should be as \emph{fast} as possible.

	\item $\mathcal{P}_{\text{Robust}}$: It should be \emph{robust} to a noisy normal field\footnote{If the normal field is estimated via photometric ste\-reo, we suppose that the images are corrupted by an additive Gaussian noise, as recommended in \cite{Noakes:2003a}: ``in previous work on photometric stereo, noise is [wrongly] added to the gradient of the height function rather than camera images''.}.
	
	\item $\mathcal{P}_{\text{FreeB}}$: The method should be able to handle a \emph{free boundary}. Accordingly, each method aimed at solving the Poisson equation \eqref{eq:23} should be able to solve the natural boundary condition \eqref{eq:31} in the same time.
	
	\item $\mathcal{P}_{\text{Disc}}$: The method should preserve the \emph{depth discontinuities}. This property could allow us, for example, to use photometric stereo on a whole image, without segmenting the scene into different parts without discontinuity.

	\item $\mathcal{P}_{\text{NoRect}}$: The method should be able to work on a \emph{non-rectangular} domain. This happens for example when photometric stereo is applied to an object with background. This property could partly remedy a method which would not satisfy $\mathcal{P}_{\text{Disc}}$, knowing that segmentation is usually easier to manage than preserving the depth discontinuities.
\end{itemizepoint}
An additional property would also be much appreciated:
\begin{itemizepoint}
	\item $\mathcal{P}_{\text{NoPar}}$: The method should have \emph{no parameter} to tune (only the \emph{critical} parameters are involved here). In practice, tuning more than one parameter often means that an expert of the method is needed. One parameter is often considered as acceptable, but no parameter is even better.
\end{itemizepoint}


\subsection{Integration and Integrability} \label{sec:4.2}

Among the required properties, we did not \emph{explicitly} quote the ability of a method to deal with non-integrable normal fields, but this is \emph{implicitly} expected through $\mathcal{P}_{\text{Robust}}$ and $\mathcal{P}_{\text{Disc}}$. In other words, the two sources of non-integrability that we consider are noise and depth discontinuities. A normal field estimated using shape-from-shading could be very far from being integrable, because of the ill-posedness of this technique, to such a point that integrability is sometimes used to disambiguate the problem \cite{Frankot:1988a,Horn:1986a}. Also the \emph{uncalibrated} photometric stereo problem is ill-posed, and is systematically disambiguated imposing integrability \cite{Yuille:1997}. However, we are over all interested in \emph{calibrated} photometric stereo with $n \geq 3$ images, which is well-posed without resorting to integrability. An error in the intensity of one light source is enough to cause a bias \cite{Horovitz:2004a}, and outliers may appear in shadow regions \cite{Reddy:2009a}, thus providing normal fields that can be highly non-integrable, but we argue that such defects do not have to be compensated by the integration method itself. In other words, we suppose that the only outliers of the normal field we want to integrate are located on depth discontinuities.

In order to know whether a normal integration me\-thod satisfies $\mathcal{P}_{\text{Disc}}$, a shape like $\mathcal{S}_{\text{vase}}$ (see Figure \ref{fig:5}-a) is well indicated, but a practical mistake must be avoided, which is not obvious. A discrete approximation of the \emph{integrability term} $\iint_{(u,v) \in \Omega} [\partial_v p(u,v) -\partial_u q(u,v)]^2 \,\mathrm{d}u\,\mathrm{d}v$, which is used in \cite{Horn:1986a} to measure the departure of a gradient field $[p,q]^\top$ from being integrable, is as follows:
\begin{equation}
	\displaystyle E_{\text{int}} = \displaystyle \sum_{(u,v) \in \Omega_3}\left[\frac{p_{u,v+1}-p_{u,v}}{\delta}-\frac{q_{u+1,v}-q_{u,v}}{\delta}\right]^2
\label{eq:53}
\end{equation}
where $\Omega_3$ denotes the set of pixels $(u,v)\in\Omega$ such that $(u,v+1)$ and $(u+1,v)$ are inside $\Omega$. Let us suppose in addition that the discrete values $p_{u,v}$ and $q_{u,v}$ are \emph{numerically approximated} using as finite differences:
\begin{equation}
	\left\{
	\begin{aligned}
		p_{u,v} = \frac{z_{u+1,v}-z_{u,v}}{\delta} \\
		q_{u,v} = \frac{z_{u,v+1}-z_{u,v}}{\delta}
	\end{aligned}
	\right.
\label{eq:54}
\end{equation}
Reporting the expressions \eqref{eq:54} of $p_{u,v}$ and $q_{u,v}$ in \eqref{eq:53}, this always implies $E_{\text{int}} = 0$. Using such a numerically approximated gradient field is thus biased, since it is integrable even in the presence of discontinuities\footnote{This problem is also noted by Saracchini et al. in \cite{Saracchini:2012a}: ``Note that taking finite differences of the reference height map will not yield adequate test data''.}, whereas for instance, $E_{\text{int}} = 390$ in the case of $\*g_{\text{vase}}$.


\subsection{Most Representative Normal Integration Methods} \label{sec:4.3}

The problem of normal integration is sometimes considered as solved, because its mathematical formulation is well established (see Section \ref{sec:2}), but we will see that none of the existing methods simultaneously satisfies all the required properties. In 1996, Klette and Schl{\"u}ns stated that ``there is a remarkable deficiency of literature about integration techniques, at least in computer vision''~\cite{Klette:1996a}. Many contributions have appeared afterwards, but a detailed review is still missing.

The way to cope with a possible non-integrable normal field was seen as a property of primary importance in the first papers on normal integration. The most obvious way to solve the problem amounts to use different paths in the integrals of \eqref{eq:5}, \eqref{eq:9} or \eqref{eq:18}, and to average the different values. Apart from this approach, which has given rise to several heuristics \cite{Coleman:1982a,Healey:1984a,Wu_Z:1988a}, we propose to separate the main existing normal integration methods into two classes, depending on whether they care about discontinuities or not.


\subsubsection{Methods which do not care about Discontinuities} \label{sec:4.3.1}

According to the discussion conducted in Section \ref{sec:2.3}, the most natural way to overcome non-integrability is to solve the Poisson equation \eqref{eq:23}. This approach has given rise to the method $\mathcal{M}_{\text{HB}}$ (see Section \ref{sec:3.1}), pioneered by Ikeuchi in \cite{Ikeuchi:1983a} and then detailed by Horn and Brooks in \cite{Horn:1986a}, which has been the source of inspiration of several subsequent works. The method $\mathcal{M}_{\text{HB}}$ satisfies the property $\mathcal{P}_{\text{Robust}}$ (much better than the heuristics cited above), as well as $\mathcal{P}_{\text{NoPar}}$ (there is no parameter), $\mathcal{P}_{\text{NoRect}}$ and $\mathcal{P}_{\text{FreeB}}$. A drawback of $\mathcal{M}_{\text{HB}}$ is that it does not satisfy $\mathcal{P}_{\text{Fast}}$, since it uses a Jacobi iteration to solve a large linear system whose size is equal to the number of pixels inside $\mathring{\Omega} = \Omega \backslash\partial\Omega$. Unsurprisingly, $\mathcal{P}_{\text{Disc}}$ is not satisfied by $\mathcal{M}_{\text{HB}}$, since this method does not care about discontinuities.

Frankot and Chellappa address shape-from-shading using a method which ``also can be used as an integrator'' \cite{Frankot:1988a}. The gradient field is projected on a set $\mathcal{I}$ of integrable vector fields $\nabla z$. In practice, the set of functions $z$ is spanned by the Fourier basis. The method $\mathcal{M}_{\text{FC}}$ (see Section \ref{sec:3.3}) not only satisfies $\mathcal{P}_{\text{Robust}}$ and $\mathcal{P}_{\text{NoPar}}$, but also $\mathcal{P}_{\text{Fast}}$, since it is non-iterative and, thanks to the FFT algorithm, much faster than $\mathcal{M}_{\text{HB}}$. On the other hand, the reconstruction domain is implicitly supposed to be rectangular, even if the following is claimed: ``The Fourier expansion could be formulated on a finite lattice instead of a periodic lattice. The mathematics are somewhat more complicated [...] and more careful attention could then be paid to boundary conditions''. Hence, $\mathcal{P}_{\text{NoRect}}$ is not really satisfied, not more than $\mathcal{P}_{\text{Disc}}$. Finally, $\mathcal{P}_{\text{FreeB}}$ is not satisfied, since the solution is constrained to be periodic, which can cause large errors\footnote{Noakes, Kozera and Klette follow the same way as Frankot and Chellappa but, since they use a set $\mathcal{I}$ of integrable vector fields which is not spanned by the Fourier basis, they cannot resort to the very efficient Fast Fourier Transform algorithm any more \cite{Noakes:2001a,Noakes:1999b}.}.

Simchony, Chellappa and Shao suggest in \cite{Simchony:1990a} a non-iterative way to solve the Poisson equation \eqref{eq:23} using direct analytical methods. It is noteworthy that the resolution of the discrete Poisson equation described in Section \ref{sec:3.4}, in the case of a Dirichlet or Neumann boundary condition, does not exactly match the original description in \cite{Simchony:1990a}, but is rather intended to be pedagogic. As observed by Lee in \cite{Lee_D:1985a}, the discretized Laplacian operator on a rectangular domain is a symmetric tridiagonal Toeplitz matrix if a Dirichlet boundary condition is used, whose eigenvalues are analytically known. Simchony, Chellappa and Shao show how to use the \emph{discrete sine transform} to diagonalize such a matrix. They design an efficient solver for Equation \eqref{eq:23}, which satisfies $\mathcal{P}_{\text{Fast}}$, $\mathcal{P}_{\text{Robust}}$ and $\mathcal{P}_{\text{NoPar}}$. Another extension of $\mathcal{M}_{\text{SCS}}$ to Neumann boundary conditions using the \emph{discrete cosine transform} is suggested, which allows $\mathcal{P}_{\text{FreeB}}$ to be satisfied. Even if \emph{embedding techniques} are supposed to generalize this method to non-rectangular domains, $\mathcal{P}_{\text{NoRect}}$ is not satisfied in practice. Neither is $\mathcal{P}_{\text{Disc}}$.

In \cite{Horovitz:2000,Horovitz:2004a}, Horovitz and Kiryati improve $\mathcal{M}_{\text{HB}}$ in two ways. They show how to incorporate the depth in some sparse \emph{control points}, in order to correct a possible bias in the reconstruction. They also design a coarse-to-fine multigrid computation, in order to satisfy $\mathcal{P}_{\text{Fast}}$\footnote{Goldman et al. do the same using the conjugate gradient method \cite{Goldman:2005a}.}. This acceleration technique however requires a parameter to be tuned, which loses $\mathcal{P}_{\text{NoPar}}$.

As in \cite{Frankot:1988a}, Petrovic et al. enforce integrability \cite{Petrovic:2001a}, but the normal field is directly handled under its discrete writing, in a Bayesian framework. ``Imposing the integrability over elementary loops in [a] graphical model will correct the irregularities in the data''. An iterative algorithm known as \emph{belief propagation} is used to converge towards the MAP estimate of the unknown surface. It is claimed that ``discontinuities are maintained'', but too few results are provided to decide whether $\mathcal{P}_{\text{Disc}}$ is really satisfied.

In \cite{Kimmel:2003}, Kimmel and Yavneh show how to accelerate the multigrid method designed by Horovitz and Kiryati \cite{Horovitz:2000} in the case where ``the surface height at specific coordinates or along a curve'' is known, using an \emph{algebraic multigrid} approach. Basically, their method has the same properties as \cite{Horovitz:2000}, although $\mathcal{P}_{\text{Fast}}$ is even better satisfied.

An alternative derivation of Equation \eqref{eq:37} is yielded by Wei and Klette in \cite{Wei_T:2003a}, in which the preliminary derivation of the Euler-Lagrange equation \eqref{eq:23} associated to $\mathcal{F}_{L_2}(z)$ is not needed. They claim that ``to solve the minimization problem, we employ the Fourier transform theory rather than variational approach to avoid using the initial and boundary conditions'', but since a periodic boundary condition is actually used instead, $\mathcal{P}_{\text{FreeB}}$ is not satisfied. They also add two regularization terms to the functional $\mathcal{F}_{L_2}(z)$ given in \eqref{eq:21}, ``in order to improve the accuracy and robustness''. The property $\mathcal{P}_{\text{NoPar}}$ is thus lost. On the other hand, even if Wei and Klette note that ``[$\mathcal{M}_{\text{HB}}$] is very sensitive to abrupt changes in orientation, i.e., there are large errors at the object boundary'', their method does not satisfy $\mathcal{P}_{\text{Disc}}$ either, whereas we will observe that loosing $\mathcal{P}_{\text{NoPar}}$ is often the price to satisfy $\mathcal{P}_{\text{Disc}}$.

Another method inspired by $\mathcal{M}_{\text{FC}}$ is that of Kovesi \cite{Kovesi:2005a}. Instead of projecting the given gradient field on a Fourier basis, Kovesi suggests to compute the correlations of this gradient field with the gradient fields of a \emph{bank of shapelets}, which are in practice a family of Gaussian surfaces\footnote{According to \cite{Agrawal:2006a}, Kovesi uses ``a redundant set of non-orthogonal basis functions''.}. This method globally satisfies the same properties as $\mathcal{M}_{\text{FC}}$ but, in addition, it can be applied to an \emph{incomplete normal field} i.e., to normals whose tilts are known up to a certain ambiguity\footnote{\url{http://www.peterkovesi.com/matlabfns/index.html\#shapelet}}. Although photometric stereo computes the normals without ambiguity, this peculiarity could indeed be useful when a couple of images only is used, or \emph{a fortiori} a single image (shape-from-shading), since the problem is not well-posed in both these cases.

In \cite{Ho:2006a}, Ho, Lim, Yang and Kriegman derive from \eqref{eq:20} the \emph{eikonal equation} $\|\nabla z\|^2 = p^2+q^2$, and aspire to use the \emph{fast marching} method for its resolution. Unfortunately, this method requires that the unknown $z$ has a unique global minimum over $\Omega$. Thereby, a more general eikonal equation $\|\nabla (z+\lambda\, f)\|^2 = (p+\lambda\, \partial_u f)^2+(q+\lambda\, \partial_v f)^2$ is solved, where $f$ is a known function and the parameter $\lambda$ has to be tuned so that $z+\lambda\, f$ has a unique global minimum. The main advantage is that $\mathcal{P}_{\text{Fast}}$ is (widely) satisfied. Nothing is said about robustness, but we guess that error accumulation occurs as the depth is computed
level set by level set.

As already explained in Section \ref{sec:3.2}, Durou and Cour\-teille improve $\mathcal{M}_{\text{HB}}$ in \cite{Durou:2007a}, in order to better satisfy $\mathcal{P}_{\text{FreeB}}$\footnote{A preliminary version of this method was already described in \cite{Crouzil:2003a}.}. A very similar improvement of $\mathcal{M}_{\text{HB}}$ is proposed by Harker and O'Leary in \cite{Harker:2008a}. The latter loses the ability to handle any reconstruction domain, whereas the reformulation of the problem as a \emph{Sylvester equation} provides two appreciable improvements. First, it deals with matrices of the same size as the initial (regular) grid and resorts to very efficient solvers dedicated to Sylvester equations, thus satisfying $\mathcal{P}_{\text{Fast}}$. Moreover, any form of discrete derivatives is allowed. In \cite{Harker:2011a,Harker:2015a}, Harker and O'Leary moreover propose several variants including regularization, which are still written as Syl\-vester equations, but one of them loses the property $\mathcal{P}_{\text{FreeB}}$, whereas the others lose $\mathcal{P}_{\text{NoPar}}$.

In \cite{Ettl:2008a}, Ettl et al. propose a method specifically designed for \emph{deflectometry}, which aims at measuring ``height variations as small as a few nanometers'' and delivers normal fields ``with small noise and curl'', but the normals are provided on an \emph{irregular grid}. This is why Ettl et al. search for an interpolating/approximating surface rather than for one unknown value per sample. Of course, this method is highly parametric, but $\mathcal{P}_{\text{NoPar}}$ is still satisfied, since the parameters are the unknowns. Its main problem is that $\mathcal{P}_{\text{Fast}}$ is rarely satisfied, depending on the number of parameters that are used.

The fast-marching method \cite{Ho:2006a} is improved in \cite{Galliani:2012a} by Galliani, Breu{\ss} and Ju in three ways. First, the method by Ho et al. is shown to be inaccurate, due to the use of analytical derivatives $\partial_u f$ and $\partial_v f$ in the eikonal equation, instead of discrete derivatives. An \emph{upwind scheme} is more appropriate to solve such a PDE. Second, the new method is more stable and the choice of $\lambda$ is no more a cause for concern. This implies that $\mathcal{P}_{\text{NoPar}}$ is satisfied \emph{de facto}. Finally, any form of domain $\Omega$ can be handled, but it is not clear whether $\mathcal{P}_{\text{NoRect}}$ was not satisfied by the former method yet. Not surprisingly, this new method is not robust, even its if robustness is improved in a more recent paper \cite{Bahr:2015a}, but it can be used as initialization for more robust methods based, for instance, on quadratic regularization \cite{Breuss:2017}.

The integration method proposed by Balzer in \cite{Balzer:2012a} is based on second order shape derivatives, allowing for the use of a fast Gauss-Newton algorithm. Hence $\mathcal{P}_{\text{Fast}}$ is satisfied. A careful meshing of the problem, as well as the use of a finite element method, make $\mathcal{P}_{\text{NoRect}}$ to be satisfied. However, for the very same reason, $\mathcal{P}_{\text{NoPar}}$ is not satisfied. Moreover, the method is limited to smooth surfaces, and therefore $\mathcal{P}_{\text{Disc}}$ cannot be satisfied. Finally, $\mathcal{P}_{\text{Robust}}$ can be achieved thanks to a premiminary filtering step. Balzer and M{\"o}rwald design in \cite{Balzer:2012b} another finite element method, where the surface model is based on \emph{B-splines}. It satisfies the same properties as the previous method\footnote{\url{https://github.com/jonabalzer/iga-integration/tree/master/core}}.

In \cite{Xie:2014a}, Xie et al. deform a mesh ``to let its facets follow the demanded normal vectors'', resorting to discrete geometry processing. As a non-parametric surface model is used, $\mathcal{P}_{\text{FreeB}}$, $\mathcal{P}_{\text{NoPar}}$ and $\mathcal{P}_{\text{NoRect}}$ are satisfied. In order to avoid the oversmoothing effect of many previous methods, sharp features can be preserved. However, $\mathcal{P}_{\text{Disc}}$ is not addressed. On the other hand, $\mathcal{P}_{\text{Fast}}$ is not satisfied since the proposed method alternates local and global optimization.

In \cite{Yamaura:2015}, Yamaura et al. design a new method based on B-splines, which has the same properties as that proposed in \cite{Balzer:2012b}. But since the latter ``relies on second-order partial differential equations, which is inefficient and unnecessary, as normal vectors consist of only first-order derivatives'', a simpler formalism with higher performances is proposed. Moreover, a nice application to surface editing is exhibited.


\subsubsection{Methods which care about Discontinuities} \label{sec:4.3.2}

The first work which \emph{really} addresses the problem of $\mathcal{P}_{\text{Disc}}$ is by Kara\c{c}ali and Snyder \cite{Karacali:2003a,Karacali:2004a}, who show how to define a new orthonormal basis of integrable vector fields which can incorporate depth discontinuities. They moreover show how to detect such discontinuities, in order to \emph{partially} enforce integrability. The designed method thus satisfies $\mathcal{P}_{\text{Disc}}$, as well as $\mathcal{P}_{\text{Robust}}$ and $\mathcal{P}_{\text{FreeB}}$, but $\mathcal{P}_{\text{Fast}}$ is lost, despite the use of a block processing technique inspired by the work of Noakes et al. \cite{Noakes:2001a,Noakes:1999b}. In accordance with a previous remark, since it is often the price to satisfy $\mathcal{P}_{\text{Disc}}$, $\mathcal{P}_{\text{NoPar}}$ is also lost.

In \cite{Agrawal:2005a}, Agrawal, Chellappa and Raskar consider the pixels as a weighted graph, such that the weights are of the form $p_{u,v+1}-p_{u,v}-q_{u+1,v}+q_{u,v}$. Each edge whose weight is greater than a threshold is cut. A minimal number of suppressed edges are then restored, in order to reconnect the graph while minimizing the total weight. As soon as an edge is still missing, one gradient value $p_{u,v}$ or $q_{u,v}$ is considered as possibly corrupted. It is shown how these suspected gradient values can be corrected, in order to enforce integrability ``with the important property of local error confinement''. Neither $\mathcal{P}_{\text{FreeB}}$ nor $\mathcal{P}_{\text{NoPar}}$ is satisfied, and $\mathcal{P}_{\text{Fast}}$ is not guaranteed as well, even if the method is non-iterative. Moreover, the following is asserted in \cite{Reddy:2009a}: ``Under noise, the algorithm in \cite{Agrawal:2005a} confuses correct gradients as outliers and performs poorly''. Finally, $\mathcal{P}_{\text{Disc}}$ may be satisfied since strict integrability is no more uniformly imposed over the entire gradient field.

In \cite{Agrawal:2006a}, Agrawal, Raskar and Chellappa propose a general framework ``based on controlling the anisotropy of weights for gradients during the integration''\footnote{\url{http://www.amitkagrawal.com/eccv06/RangeofSurfaceReconstructions.html}}. They are much inspired by classical image restoration techniques. It is shown how the (isotropic) Laplacian operator in Equation \eqref{eq:23} must be modified using ``spatially varying anisotropic kernels'', thus obtaining four methods: two based on robust estimation, one on regularization, and one on anisotropic diffusion\footnote{A similar approach will be detailed in our second paper.}. A homogeneous Neumann boundary condition $\nabla z \cdot \boldsymbol{\eta} = 0$ is assumed, which looks rather unrealistic. Thus, the four proposed methods do not satisfy $\mathcal{P}_{\text{FreeB}}$, neither $\mathcal{P}_{\text{Fast}}$ nor $\mathcal{P}_{\text{NoPar}}$. Nevertheless, a special attention is given to satisfy $\mathcal{P}_{\text{Robust}}$ and $\mathcal{P}_{\text{Disc}}$.

A similar method to the first one proposed in \cite{Agrawal:2006a} is designed by Fraile and Hancock in \cite{Fraile:2006a}. A \emph{minimum spanning tree} is constructed from the same graph of pixels as in \cite{Agrawal:2005a}, except that the weights are different (several weights are tested). The integral in \eqref{eq:6} along the unique path joining each pixel to a root pixel is then computed. Of course, this method is less robust than those based on quadratic regularization (or than the weighted quadratic regularization proposed in \cite{Agrawal:2006a}), since ``the error due to measurement noise propagates along the path'', and it is rather slow because of the search for a minimum spanning tree. But, depending on which weights are used, it could preserve depth discontinuities: in such a case as that of Figure \ref{fig:5}-a, each pixel could be reached from a root pixel without crossing any discontinuity.

In \cite{Wu_TP:2006b}, Wu and Tang try to find the best compromise between integrability and discontinuity preservation. In order to segment the scene into pieces without discontinuities, ``one plausible method [...] is to identify where the integrability constraint is violated'', but ``in real case, [this] may produce very poor discontinuity maps rendering them unusable at all''. A probabilistic method using the EM (Expectation-Maximization) algorithm is thus proposed, which provides a weighted discontinuity map. The alternating iterative optimization is very slow and a parameter is used, but this approach is promising, even if the evaluation of the results remains qualitative.

In \cite{Ng_HS:2007a,Ng_HS:2010a}, Ng, Wu and Tang do not enforce integrability over the entire domain, because with such an enforcement ``sharp features will be smoothed out and surface distortion will be produced''. Since ``either sparse or dense, residing on a 2D regular (image) or irregular grid space'' gradient fields may be integrated, it is concluded that ``a continuous formulation for surface-from-gradients is preferred''. Gaussian \emph{kernel functions} are used, in order to linearize the problem and to avoid the need for extra knowledge on the boundary\footnote{\url{http://www.cse.ust.hk/~pang/papers/supp\_materials/pami\_sur3d\_code.zip}}. Unfortunately, at least two parameters must be tuned. Also $\mathcal{P}_{\text{Fast}}$ is not satisfied, not more than $\mathcal{P}_{\text{FreeB}}$, as shown in \cite{Balzer:2012b} on the basis of several examples. On the other hand, the proposed method outputs ``continuous 3D representation not limited to a height field''.

In \cite{Reddy:2009a}, Reddy, Agrawal and Chellappa propose a method specifically designed to handle heavily corrupted gradient fields, which combines ``the best of least squares and combinatorial search to handle noise and correct outliers respectively''. Even if it is claimed that ``$L_1$ solution performs well across all scenarios without the need for any tunable parameter adjustments'', $\mathcal{P}_{\text{NoPar}}$ is not satisfied in practice. Neither is $\mathcal{P}_{\text{FreeB}}$.

In \cite{Durou:2009a}, Durou, Aujol and Courteille are mainly concerned by $\mathcal{P}_{\text{Disc}}$. Knowing that quadratic regularization works well in the case of smooth surfaces, but is not well adapted to discontinuities, the use of other regularizers, or of other variational models inspired by image processing, as in \cite{Agrawal:2006a}, allows $\mathcal{P}_{\text{Disc}}$ to be satisfied. This is detailed in the companion paper \cite{Queau:2017}.

The integration method proposed in \cite{Saracchini:2010a,Saracchini:2012a} by Saracchini, Stolfi, Leit{\~a}o, Atkinson and Smith\footnote{\url{http://www.ic.unicamp.br/stolfi/EXPORT/projects/photo-stereo/}} is a multi-scale version of $\mathcal{M}_{\text{HB}}$. A system is solved at each scale using a Gauss-Seidel iteration in order to satisfy $\mathcal{P}_{\text{Fast}}$. Since reliability in the gradient is used as local weight, ``each equation can be tuned to ignore bad data samples and suspected discontinuities'', thus allowing $\mathcal{P}_{\text{Disc}}$ to be satisfied. Finally, setting the weights outside $\Omega$ to zero allows $\mathcal{P}_{\text{FreeB}}$ and $\mathcal{P}_{\text{NoRect}}$ to be satisfied as well. However, this method ``assumes that the slope and weight maps are given''. But such a weight map is a crucial clue, and the following assertion somehow avoids the problem: ``practical integration algorithms require the user to provide a weight map''.

A similar approach is followed by Wang, Bu, Li, Song and Tan in \cite{Wang:2012a}, but the weight map is \emph{binary} and automatically computed. In addition to the gradient map, the photometric stereo images themselves are required. Eight cues are used by two SVM classifiers, which have to be trained using synthetic labelled data. Even if the results are nice, the proposed method seems rather difficult to manage in practice, and clearly loses $\mathcal{P}_{\text{Fast}}$ and $\mathcal{P}_{\text{NoPar}}$.

In \cite{Badri:2016a,Badri:2014a}, Badri, Yahia and Aboutajdine resort to $L_p$ norms, $p\in\,]0,1[$. As $p$ decreases, the $L_0$ norm is approximated, which is a sparse estimator well adapted to outliers. Indeed, the combination of four terms allows Badri et al. to design a method which simultaneously handles noise and outliers, thus ensuring that $\mathcal{P}_{\text{Robust}}$ and $\mathcal{P}_{\text{Disc}}$ are satisfied. However, since the problem becomes non-convex, the proposed \emph{half-quad} resolution is iterative and requires a good initialization: $\mathcal{P}_{\text{Fast}}$ is lost. It happens that neither $\mathcal{P}_{\text{NoPar}}$ is satisfied. Finally, each iteration resorts to FFT, which implies a rectangular domain and a periodic boundary condition, hence $\mathcal{P}_{\text{FreeB}}$ and $\mathcal{P}_{\text{NoRect}}$ are not satisfied either.


\subsection{Summary of the Review} \label{sec:4.4}

Our discussion on the most representative methods of integration is summarized in Table \ref{tab:1}, where the methods are listed in chronological order. It appears that none of them satisfies all the required properties, which is not surprising. Moreover, even if accuracy is the most basic property of any 3D-reconstruction technique, let us recall that it would have been impossible in practice to numerically compare all these methods.

\begin{table*}[htbp]
\caption[]{Main methods of integration listed in chronological order.}
\label{tab:1}
\begin{center}
{
\begin{tabular}{|c|c|c|c|c|c|c|c|c|}
	\hline
	Authors &~Year~&~Ref.~& $\mathcal{P}_{\text{Fast}}$ & $\mathcal{P}_{\text{Robust}}$ & $\mathcal{P}_{\text{FreeB}}$ & $\mathcal{P}_{\text{Disc}}$ & $\mathcal{P}_{\text{NoPar}}$ & $\mathcal{P}_{\text{NoRect}}$ \\
	\hline
	Coleman and Jain & 1982 & \cite{Coleman:1982a} & $+$ & $-$ & $+$ & $-$ & $+$ & $+$ \\
	Horn and Brooks & 1986 & \cite{Horn:1986a} & $-$ & $+$ & $+$ & $-$ & $+$ & $+$ \\
	Frankot and Chellappa & 1988 & \cite{Frankot:1988a} & $+$ & $+$ & $-$ & $-$ & $+$ & $-$ \\
	Simchony, Chellappa and Shao & 1990 & \cite{Simchony:1990a} & $+$ & $+$ & $+$ & $-$ & $+$ & $-$ \\
	Noakes, Kozera and Klette & 1999 & \cite{Noakes:1999b} & $-$ & $+$ & $-$ & $-$ & $+$ & $-$ \\
	Horovitz and Kiryati & 2000 & \cite{Horovitz:2000} & $+$ & $+$ & $+$ & $-$ & $-$ & $+$ \\
	Petrovic et al. & 2001 & \cite{Petrovic:2001a} & $-$ & $+$ & $+$ & $-$ & $-$ & $+$ \\
	Kimmel and Yavneh & 2003 & \cite{Kimmel:2003} & $+$ & $+$ & $+$ & $-$ & $-$ & $+$ \\
	Wei and Klette & 2003 & \cite{Wei_T:2003a} & $+$ & $+$ & $-$ & $-$ & $-$ & $-$ \\
	Kara\c{c}ali and Snyder & 2003 & \cite{Karacali:2003a} & $-$ & $+$ & $+$ & $+$ & $-$ & $-$ \\
	Kovesi & 2005 & \cite{Kovesi:2005a} & $+$ & $+$ & $-$ & $-$ & $+$ & $-$ \\
	Agrawal, Chellappa and Raskar & 2005 & \cite{Agrawal:2005a} & $-$ & $-$ & $-$ & $+$ & $-$ & $+$ \\
	Agrawal, Raskar and Chellappa & 2006 & \cite{Agrawal:2006a} & $-$ & $+$ & $-$ & $+$ & $-$ & $+$ \\
	Fraile and Hancock & 2006 & \cite{Fraile:2006a} & $-$ & $-$ & $+$ & $+$ & $+$ & $+$ \\
	Ho, Lim, Yang and Kriegman & 2006 & \cite{Ho:2006a} & $+$ & $-$ & $+$ & $-$ & $-$ & $+$ \\
	Wu and Tang & 2006 & \cite{Wu_TP:2006b} & $-$ & $+$ & $+$ & $+$ & $-$ & $+$ \\
	Ng, Wu and Tang & 2007 & \cite{Ng_HS:2007a} & $-$ & $+$ & $-$ & $+$ & $-$ & $+$ \\
	Durou and Courteille & 2007 & \cite{Durou:2007a} & $-$ & $+$ & $+$ & $-$ & $+$ & $+$ \\
	Harker and O'Leary & 2008 & \cite{Harker:2008a} & $+$ & $+$ & $+$ & $-$ & $+$ & $-$ \\
	Ettl, Kaminski, Knauer and H{\"a}usler & 2008 & \cite{Ettl:2008a} & $-$ & $+$ & $+$ & $-$ & $+$ & $+$ \\
	Reddy, Agrawal and Chellappa & 2009 & \cite{Reddy:2009a} & $-$ & $+$ & $-$ & $+$ & $-$ & $+$ \\
	Durou, Aujol and Courteille & 2009 & \cite{Durou:2009a} & $-$ & $+$ & $+$ & $+$ & $-$ & $+$ \\
	Saracchini et al. & 2010 & \cite{Saracchini:2010a} & $+$ & $+$ & $+$ & $+$ & $-$ & $+$ \\
	Galliani, Breu{\ss} and Ju & 2012 & \cite{Galliani:2012a} & $+$ & $-$ & $+$ & $-$ & $+$ & $+$ \\
	Balzer & 2012 & \cite{Balzer:2012a} & $+$ & $+$ & $+$ & $-$ & $-$ & $+$ \\
	Wang, Bu, Li, Song and Tan & 2012 & \cite{Wang:2012a} & $-$ & $+$ & $+$ & $+$ & $-$ & $+$ \\
	Balzer and M{\"o}rwald & 2012 & \cite{Balzer:2012b} & $+$ & $+$ & $+$ & $-$ & $-$ & $+$ \\
	Xie, Zhang, Wang and Chung & 2014 & \cite{Xie:2014a} & $-$ & $+$ & $+$ & $-$ & $+$ & $+$ \\
	Badri, Yahia and Aboutajdine & 2014 & \cite{Badri:2014a} & $-$ & $+$ & $-$ & $+$ & $-$ & $-$ \\
	Yamaura, Nanya, Imoto and Maekawa & 2015 & \cite{Yamaura:2015} & $+$ & $+$ & $+$ & $-$ & $-$ & $+$ \\
	B\"ahr et al. & 2016 & \cite{Breuss:2017} & $+$ & $+$ & $+$ & $-$ & $+$ & $+$ \\
	\hline
\end{tabular}
}
\end{center}
\end{table*}

In view of Table \ref{tab:1}, almost every method differs from all the others, regarding the six selected properties. Of course, it may appear that such a binary ($+$/$-$) table is hardly informative, but more levels in each criterion would have led to more arbitrary scores. On the other hand, the number of $+$ should not be considered as a global score for a given method: it happens that a method perfectly satisfies a subset of properties, while it does not care at all about the others.


\section{Conclusion and Perspectives} \label{sec:5}

Even if robustness to outliers was not selected in our list of required properties, let us cite a paper specifically dedicated to this problem. In \cite{Du:2007a}, Du, Robles-Kelly and Lu compare the $L_2$ and $L_1$ norms, as well as a number of M-estimators, faced to the presence of outliers in the normal field: $L_1$ is shown to be the globally best parry. This paper being worthwhile, one can wonder why it does not appear in Table \ref{tab:1}. On the one hand, even if $\mathcal{P}_{\text{Robust}}$ is satisfied at best, none of the other criteria are considered. On the other hand, let us recall why we did not select robustness to outliers as a pertinent feature: the presence of outliers in the normal field does not have to be compensated by the integration method. In \cite{Badri:2014a}, it is said that ``[photometric stereo] can fail due to the presence of shadows and noise'', but recall that photometric stereo can be robust to outliers \cite{Ikehata:2014a,Wu_L:2010}.

Another property was ignored: whether the depth can be fixed at some points or not. As integration is a well-posed problem without any additional knowledge on the solution, we considered that this property is appreciable, although not required. Let us however quote, once again, the papers by Horovitz and Kiryati \cite{Horovitz:2004a} and by Kimmel and Yavneh \cite{Kimmel:2003}, in which this problem is specifically dealt with.

Some other works on normal integration have not been mentioned in our review, since they address other problems or do not face the same challenges. Let us first cite a work by Balzer \cite{Balzer:2011a}, in which a specific problem with the normals delivered by deflectometry is highlighted: as noted by Ettl et al. in \cite{Ettl:2008a}, such normals are usually not noisy, but Balzer points out that they are \emph{distant-dependent}, which means that the gradient field $\*g = [p, q]^\top$ also depends on the depth $z$. The iterative method proposed in \cite{Balzer:2011a} to solve this more general problem seems to be quite limited, but an interesting extension to normals provided by photometric stereo is suggested: ``one could abandon the widespread assumption that the light sources are distant and the lighting directions thus constant''.

Finally, Chang et al. address in \cite{Chang:2007a} the problem of \emph{multiview normal integration}, which aims at reconstructing a full 3D-shape in the framework of multiview photometric stereo. The original variational formulation \eqref{eq:21} is extended to such normal fields, and the resulting PDE is solved via a level set method, which has to be soundly initialized. The results are nice but are limited to synthetic multiview photometric stereo images. However, this approach should be continued, since it provides complete 3D-models. Moreover, it is noted in \cite{Chang:2007a} that the use of multiview inputs is the most intuitive way to satisfy $\mathcal{P}_{\text{Disc}}$.

As noticed by Agrawal et al. about the range of solutions proposed in \cite{Agrawal:2006a}, but this is more generally true, ``the choice of using a particular algorithm for a given application remains an open problem''. We hope this review will help the reader to make up its own mind, faced to so many existing approaches.

Finally, even if none of the reviewed methods satisfies all the selected criteria, this work helped us to develop some new normal integration methods. In a companion paper entitled \emph{Variational Methods for Normal Integration} \cite{Queau:2017}, we particularly focus on the problem of normal integration in the presence of discontinuities, which occurs as soon as there are occlusions.


\bibliographystyle{spmpsci}
\bibliography{biblio}

\end{document}